\documentclass[12pt]{article}
\usepackage{setspace}

\usepackage{enumitem}
\usepackage{natbib}
\bibpunct[:]{(}{)}{;}{a}{,}{,}

\usepackage{graphicx,color}

\usepackage{url}

\usepackage{lscape}
\usepackage{wrapfig}
\usepackage{tikz}
\usepackage{float}
\usepackage{multirow}
\usepackage{amsmath, amssymb, amsfonts, amsthm}
\usepackage{physics}
\usepackage{bm}
\usepackage{dsfont}
\renewcommand{\vec}[1]{\mathbf{#1}} 
\newcommand{\mat}[1]{\mathbf{#1}} 
\newcommand{\trans}[1]{#1^\mathsf{T}} 

\newcommand{\argmax}{\mathop{\rm arg~max}\limits}

\theoremstyle{definition}
\newtheorem{theorem}{Theorem}
\newtheorem*{theorem*}{Theorem}
\newtheorem{definition}{Definition}
\newtheorem*{definition*}{Definition}
\newtheorem{lemma}{Lemma}
\newtheorem*{lemma*}{Lemma}
\newtheorem{corollary}{Corollary}
\newtheorem*{corollary*}{Corollary}

\numberwithin{equation}{section}

\usepackage[noend]{algpseudocode}
\usepackage{algorithm}


\makeatletter
\newcommand{\subsubsubsection}{\@startsection{paragraph}{4}{\z@}%
{1.5\baselineskip \@plus.5\dp0 \@minus.2\dp0}%
{.5\baselineskip \@plus2.3\dp0}%
{\reset@font\normalsize\bfseries}
}
\newcommand{\subsubsubsubsection}{\@startsection{subparagraph}{5}{\z@}%
{1.5\baselineskip \@plus.5\dp0 \@minus.2\dp0}%
{.5\baselineskip \@plus2.3\dp0}%
{\reset@font\normalsize\itshape}
}
\makeatother
\usepackage[subrefformat=parens]{subcaption}
\usepackage[subrefformat=parens]{caption}

\makeatletter
\def\eqnarray{\stepcounter {equation}\let \@currentlabel =\theequation
\global \@eqnswtrue
\global \@eqcnt \z@ \tabskip \@centering \let \\=\@eqncr
$$\halign to \displaywidth \bgroup \@eqnsel \hskip \@centering
$\displaystyle \tabskip \z@ {##}$&\global \@eqcnt \@ne \hfil
${\mbox{}##\mbox{}}$\hfil &\global \@eqcnt \tw@
$\displaystyle \tabskip \z@ {##}$\hfil \tabskip \@centering
&\llap {##}\tabskip \z@ \cr}
\makeatother

\begin{document}

\baselineskip = 8mm

\begin{center}
\textbf{\LARGE Multilinear Common Component Analysis via Kronecker Product Representation} 
\end{center}

\begin{center}
{\large Kohei Yoshikawa$^{1}$, \ Shuichi Kawano$^{1}$}
\end{center}

\begin{center}
\begin{minipage}{14cm}
{
\begin{center}
{\it {\footnotesize

\vspace{1.2mm}

$^1$ Graduate School of Informatics and Engineering,  The University of Electro-Communications,
1-5-1 Chofugaoka, Chofu-shi, Tokyo 182-8585, Japan. \\

\vspace{1.2mm}
}}
\vspace{2mm}

yoshikawa@ai.lab.uec.ac.jp \hspace{5mm} skawano@ai.lab.uec.ac.jp \\

\end{center}


}
\end{minipage}
\end{center}

\vspace{1mm}
\begin{abstract}
\noindent We consider the problem of extracting a common structure from multiple tensor datasets.
For this purpose, we propose multilinear common component analysis (MCCA) based on Kronecker products of mode-wise covariance matrices.
MCCA constructs a common basis represented by linear combinations of the original variables which loses as little information of the multiple tensor datasets.
We also develop an estimation algorithm for MCCA that guarantees mode-wise global convergence.
Numerical studies are conducted to show the effectiveness of MCCA.

\end{abstract}
\begin{center}
\begin{minipage}{14cm}{
\vspace{3mm}
}
\end{minipage}
\end{center}

\baselineskip = 8mm


\section{Introduction}

Various statistical methodologies for extracting useful information from a large amount of data have been studied over the decades since the appearance of big data.
In the present era, it is important to discover a common structure of multiple datasets.
In an early study, \citet{Flury_1984} focused on the structure of the covariance matrices of multiple datasets and discussed the heterogeneity of the structure.
The author reported that population covariance matrices differ between multiple datasets in practical applications.
Many methodologies have been developed for treating the heterogeneity between covariance matrices of multiple datasets (see, e.g., \citet{Flury_1986, Flury_1988, Flury_1986_SIAM, Pourahmadi_2007, Wang_2011, Park_2018}).

Among such methodologies, common component analysis (CCA) \citep{Wang_2011} is an effective tool for statistics.
The central idea of CCA is to reduce the number of dimensions of data while losing as little information of the multiple datasets as possible.
To reduce the number of dimensions, CCA reconstructs the data with a few new variables which are linear combinations of the original variables.
For considering the heterogeneity between covariance matrices of multiple datasets, CCA assumes that there is a different covariance matrix for each dataset.
There have been many papers on various statistical methodologies using multiple covariance matrices: discriminant analysis \citep{Bensmail_1996}, spectral decomposition \citep{Boik_2002}, and a likelihood ratio test for multiple covariance matrices \citep{Manly_1987}.
It should be noted that principal component analysis (PCA) \citep{Pearson_1901, Jolliffe_Book} is a technique similar to CCA. In fact, CCA is a generalization of PCA; PCA can only be applied to one dataset, whereas CCA can be applied to multiple datasets.

Meanwhile, in various fields of research, including machine learning and computer vision, the main interest has been in tensor data, which has a multidimensional array structure.
In order to apply the conventional statistical methodologies, such as PCA, to tensor data, a simple approach is to first transform the tensor data into vector data and then apply the methodology.
However, such an approach causes the following problems:
\begin{enumerate}
    \item In losing the tensor structure of the data, the approach ignores the higher-order inherent relationships of the original tensor data.
    \item Transforming tensor data to vector data increases the number of features large. It also has a high computational cost.
\end{enumerate}
To overcome these problems, statistical methodologies for tensor data analyses have been proposed which take the tensor structure of the data into consideration.
Such methods enable us to accurately extract higher-order inherent relationships in a tensor dataset.
In particular, many existing statistical methodologies have been extended for tensor data, for example, multilinear principal component analysis (MPCA) \citep{Lu_2008} and sparse PCA for tensor data analysis \citep{Allen_2012, Wang_2012, Lai_2014}, as well as others (see \citet{Carroll_1970}, \citet{Harshman_1970}, \citet{Kiers_2000}, \citet{Badeau_2008}, and \citet{Kolda_2009}).

In this paper, we extend CCA to tensor data analysis, proposing \textit{multilinear common component analysis} (MCCA).
MCCA discovers the common structure of multiple datasets of tensor data while losing as little of the information of the datasets as possible.
To identify the common structure, we estimate a common basis constructed as linear combinations of the original variables.
For estimating the common basis, we develop a new estimation algorithm based on the idea of CCA.
In developing the estimation algorithm, two issues must be addressed.
One is the convergence properties of the algorithm.
The other is its computational cost.
To determine the convergence properties, we investigated first the relationship between the initial values of the parameters and global optimal solution and then the monotonic convergence of the estimation algorithm.
These analyses revealed that our proposed algorithm guarantees convergence of the mode-wise global optimal solution under some conditions.
To analyze the computational efficacy, we calculate the computational cost of our proposed algorithm and compare it with the computational cost of MPCA.

The rest of the paper is organized as follows.
In Section 2, we review the formulation and the minimization problem of CCA.
In Section 3, we formulate the MCCA model by constructing the covariance matrices of tensor data, based on a Kronecker product representation.
Then, we formulate the estimation algorithm for MCCA in Section 4.
In Section 5, we present the theoretical properties for our proposed algorithm and analyze the computational cost.
The efficacy of the MCCA is demonstrated through the results of numerical experiments in Section 6.
Concluding remarks are presented in Section 7.
Technical proofs are provided in the Appendices. 
Our implementation of MCCA and supplementary materials are available at \url{https://github.com/yoshikawa-kohei/MCCA}.

\section{Common Component Analysis}
Suppose that we obtain data matrices $\mat{X}_{(g)} = \trans{[\vec{x}_{(g)1}, \dots \vec{x}_{(g)N_g}]} \in \mathbb{R}^{N_g \times P}$ with $N_g$ observations and $P$ variables for $g=1, \dots G$, where $\vec{x}_{(g)i}$ is the $P$-dimensional vector corresponding to the $i$-th row of $\mat{X}_{(g)}$ and $G$ is the number of datasets.
Then, the sample covariance matrix in group $g$ is
\begin{align}
    \mat{S}_{(g)} = \frac{1}{N_g} \sum_{i=1}^{N_g} \qty( \vec{x}_{(g)i} - \bar{\vec{x}}_{(g)}) \qty( \vec{x}_{(g)i} - \bar{\vec{x}}_{(g)} )^\top, \quad g=1,\dots,G,
\end{align}
where $\mat{S}_{(g)} \in \mathbb{S}_+^{P}$, in which $\mathbb{S}_+^{P}$ is a set of symmetric positive definite matrices of  the size $P \times P$, and $\bar{\vec{x}}_{(g)} = \frac{1}{N_g}\sum_{i=1}^{N_g} \vec{x}_{(g)i}$ is a $P$-dimensional mean vector in group $g$.

The main idea of the CCA model is to find the common structure of multiple datasets by projecting the data onto a common lower-dimensional space with the same basis as the datasets.
\citet{Wang_2011} assumed that the covariance matrices $\mat{S}_{(g)}$ for $g = 1,\dots, G$ can be decomposed to a product of latent covariance matrices and an orthogonal matrix for the linear transformation as follows:
\begin{align}
    \label{eq:original_CCA_model}
    \mat{S}_{(g)} = \mat{V} \mat{\Lambda}_{(g)} \mat{V}^\top + \mat{E}_{(g)},\qq{s.t.} \mat{V}^\top \mat{V} = \mat{I}_R,
\end{align}
where $\mat{\Lambda}_{(g)} \in \mathbb{S}_+^{R}$ is the latent covariance matrix in group $g$, $\mat{V} \in \mathbb{R}^{P \times R}$ is an orthogonal matrix for the linear transformation, $\mat{E}_{(g)} \in\mathbb{S}_+^{P}$ is the error matrix in group $g$, and $\mat{I}_R$ is the identity matrix of size $R \times R$.
$\mat{E}_{(g)}$ consists of the sum of outer products for independent random vectors $\sum_{i=1}^{N_g} \mat{e}_{(g)i} \mat{e}_{(g)i}^\top$ with mean $\mathrm{E} \qty[\mat{e}_{(g)i}] = \bm{0}$ and covariance matrix $\mathrm{Cov} \qty[\mat{e}_{(g)i}] \ (> \bm{O}) \ (i = 1,2,\dots, N_g)$. 
$\mat{V}$ determines the $R$-dimensional common subspace of the multiple datasets.
In particular, by assuming $R < P$, the CCA can discover the latent structures of the datasets.
\citet{Wang_2011} referred to the model (\ref{eq:original_CCA_model}) as \textit{common component analysis} (CCA).

The parameters $\mat{V}$ and $\mat{\Lambda}_{(g)} \ (g=1,\ldots,G)$ are estimated by solving the minimization problem
\begin{align}
    \label{eq:CCA_min_problem}
    \min_{\substack{\mat{V}, \mat{\Lambda}_{(g)}\\g=1,\dots,G}} \sum_{g=1}^G \norm{\mat{S}_{(g)} - \mat{V} \mat{\Lambda}_{(g)} \trans{\mat{V}}}_F^2, \qq{s.t.} \mat{V}^\top \mat{V} = \mat{I}_R,
\end{align}
where $\norm{\cdot}_F$ denotes the Frobenius norm.
The estimator of latent covariance matrices $\mat{\Lambda}_{(g)}$ for $g = 1,\dots, G$ can be obtained by solving the minimization problem as $\hat{\mat{\Lambda}}_{(g)} = \trans{\mat{V}} \mat{S}_{(g)} \mat{V}$.
By using the estimated value $\hat{\mat{\Lambda}}_{(g)}$, the minimization problem can be reformulated as the following maximization problem:
\begin{align}
    \label{eq:CCA_max_problem}
    \max_{\mat{V}}  \tr \qty{ \mat{V}^\top \sum_{g=1}^G  \qty( \mat{S}_{(g)} \mat{V} \mat{V}^\top \mat{S}_{(g)} ) \mat{V} },\qq{s.t.} \mat{V}^\top \mat{V} = \mat{I}_R,
\end{align}
where $\tr (\cdot)$ denotes the trace of a matrix.
A crucial issue for solving the maximization problem (\ref{eq:CCA_max_problem}) is the non-convexity.
Certainly, the maximization problem is non-convex since the problem is defined on a set of orthogonal matrices, which is a non-convex set.
Generally speaking, it is difficult to find the global optimal solution in non-convex optimization problems, such as the problem (\ref{eq:CCA_max_problem}).
To overcome this drawback, \citet{Wang_2011} proposed an estimation algorithm in which the estimated parameters are guaranteed to constitute the global optimal solution under some conditions.

\section{Multilinear Common Component Analysis}
In this section, we introduce a mathematical formulation of the MCCA, which is an extension of the CCA in terms of tensor data analysis.
Moreover, we formulate an optimization problem of MCCA and investigate its convergence properties.

Suppose that we independently obtain $M$-th order tensor data $\mathcal{X}_{(g)i} \in \mathbb{R}^{P_1 \times P_2 \times \dots \times P_M}$ for $i=1,\dots N_g$.
We set the datasets of the tensors $\mathcal{X}_{(g)} = [\mathcal{X}_{(g)1}, \mathcal{X}_{(g)2}, \dots, \mathcal{X}_{(g)N_g}] \in \mathbb{R}^{P_1 \times P_2 \times \dots \times P_M \times N_g}$ for $g=1,\dots,G$, where $G$ is the number of datasets.
Then, the sample covariance matrix in group $g$ for the tensor dataset is defined by
\begin{align}
    \label{eq:covariance_matrix}
    \mat{S}_{(g)}^* := \mat{S}_{(g)}^{(1)} \otimes \mat{S}_{(g)}^{(2)} \otimes \cdots \otimes \mat{S}_{(g)}^{(M)},
\end{align}
where $\mat{S}_{(g)}^* \in \mathbb{S}_+^{P}$, in which $P = \prod_{k=1}^M P_k$, $\otimes$ denotes the Kronecker product operator, and $\mat{S}_{(g)}^{(k)} \in \mathbb{S}_+^{P_k}$ is the sample covariance matrix for $k$-th mode in group $g$ defined by
\begin{align}
    \label{eq:covariance_matrix_each_mode}
    \mat{S}_{(g)}^{(k)} := \frac{1}{N_g \prod_{j \neq k} P_j} \sum_{i=1}^{N_g} \qty(\mat{X}_{(g)i}^{(k)} - \bar{\mat{X}}_{(g)}^{(k)} )  \qty(\mat{X}_{(g)i}^{(k)} - \bar{\mat{X}}_{(g)}^{(k)} )^\top.
\end{align}
Here, $\mat{X}_{(g)i}^{(k)} \in \mathbb{R}^{P_k \times (\prod_{j \neq k} P_j)}$ is the mode-$k$ unfolded matrix of $\mathcal{X}_{(g)i}$ and $\bar{\mat{X}}_{(g)}^{(k)} \in \mathbb{R}^{P_k \times (\prod_{j \neq k} P_j)}$ is the mode-$k$ unfolded matrix of $\bar{\mathcal{X}}_{(g)} = \frac{1}{N_g}\sum_{i=1}^{N_g} \mathcal{X}_{(g)i}$.
Note that the mode-$k$ unfolding from an $M$-th order tensor $\mathcal{X} \in  \mathbb{R}^{P_1 \times P_2 \times \dots \times P_M}$ to a matrix $\mat{X}^{(k)} \in \mathbb{R}^{P_k \times (\prod_{j \neq k} P_j)}$ means that the tensor element $(p_1, p_2, \dots, p_M)$ maps to matrix element $(p_k, l)$, where $l=1 + \sum_{t=1, t\neq k}^M (p_t - 1) L_t$ with $L_t = \prod_{m=1, m\neq k}^{t-1} P_m$, in which $p_1, p_2, \dots, p_M$ denote the indices of the $M$-th order tensor $\mathcal{X}$.
For a more detailed description of tensor operations, see \citet{Kolda_2009}.
A representation of the tensor covariance matrix by Kronecker products is often used \citep{Kermoal_2002,Yu_2004, Werner_2008}.

To formulate CCA in terms of tensor data analysis, we consider CCA for the $k$-th mode covariance matrix in group $g$ as follows:
\begin{align}
    \label{eq:tensor_CCA_model_for_mode_k}
    \mat{S}_{(g)}^{(k)} = \mat{V}^{(k)} \mat{\Lambda}_{(g)}^{(k)} {\mat{V}^{(k)}}^\top + \mat{E}_{(g)}^{(k)}, \qq{s.t.} {\mat{V}^{(k)}}^\top \mat{V}^{(k)} = \mat{I}_{R_k},
\end{align}
where $\mat{\Lambda}_{(g)}^{(k)} \in \mathbb{S}_+^{{R}_k}$ is the latent $k$-th mode covariance matrix in group $g$, $\mat{V}^{(k)} \in \mathbb{R}^{P_k \times R_k}$ is an orthogonal matrix for the linear transformation, and $\mat{E}_{(g)}^{(k)} \in \mathbb{S}_+^{{P}_k}$ is the error matrix in group $g$.
$\mat{E}_{(g)}^{(k)}$ consists of the sum of outer products for independent random vectors $\sum_{i=1}^{N_g} \mat{e}_{(g)i}^{(k)}{\mat{e}_{(g)i}^{(k)}}^\top$ with mean $\mathrm{E} \qty[\mat{e}_{(g)i}^{(k)}] = \bm{0}$ and covariance matrix $\mathrm{Cov} \qty[\mat{e}_{(g)i}^{(k)}] \ (>\bm{O}) \ (i = 1,2,\dots, N_g)$. 
Since $\mat{S}_{(g)}^*$ can be decomposed to a Kronecker product of $\mat{S}_{(g)}^{(k)}$ for $k=1, \dots, M$ in the formula (\ref{eq:covariance_matrix}), we obtain the following model:
\begin{align}
    \label{eq:tensor_CCA_model}
    \mat{S}_{(g)}^* = \mat{V}^* \mat{\Lambda}_{(g)}^* \trans{{\mat{V}^*}} + \mat{E}_{(g)}^*, \qq{s.t.} \trans{{\mat{V}^*}} \mat{V}^* = \mat{I}_R,
\end{align}
where $R=\prod_{k=1}^M R_k$, $\mat{V}^* = \mat{V}^{(1)} \otimes \mat{V}^{(2)} \otimes \cdots \otimes \mat{V}^{(M)}$, $\mat{\Lambda}_{(g)}^* = \mat{\Lambda}_{(g)}^{(1)} \otimes \mat{\Lambda}_{(g)}^{(2)} \otimes \cdots \otimes \mat{\Lambda}_{(g)}^{(M)}$, and $\mat{E}_{(g)}^*$ is the error matrix in group $g$.
We refer to this model as \textit{multilinear common component analysis} (MCCA).

To find the $R$-dimensional common subspace between the multiple tensor datasets, MCCA determines $\mat{V}^{(1)}, \mat{V}^{(2)}, \dots, \mat{V}^{(M)}$.
As with CCA, we obtain the estimate of $\mat{\Lambda}_{(g)}^*$ for $g = 1, \dots, G$ as $\hat{\mat{\Lambda}}_{(g)}^* = {\mat{V}^*}^\top \mat{S}_{(g)}^* \mat{V}^*$.
With respect to $\mat{V}^*$, we can obtain the estimate by solving the following maximization problem, which is similar to (\ref{eq:CCA_max_problem}):
\begin{align}
    \label{eq:CCA_max_problem_rplc_ast}
    \max_{\mat{V}^*}  \tr \qty{ \trans{{\mat{V}^*}} \sum_{g=1}^G  \qty( \mat{S}_{(g)}^* \mat{V}^* \trans{{\mat{V}^*}} \mat{S}_{(g)}^* ) \mat{V}^* },\qq{s.t.} \trans{{\mat{V}^*}} \mat{V}^* = \mat{I}_R,
\end{align}
However, the number of parameters will be very large when we try to solve this problem directly.
This large number of parameters result in a high computational cost.
Moreover, it may not be possible to discover the inherent relationships between the variables in each mode simply by solving the problem (\ref{eq:CCA_max_problem_rplc_ast}).

To solve the maximization problem efficiently and identify the inherent relationships, the maximization problem (\ref{eq:CCA_max_problem_rplc_ast}) can be decomposed into the mode-wise maximization problems represented in the following lemma.
\begin{lemma}
    \label{lemma:MCCA_reformulate_max_problem}
    \textit{
    An estimate of the parameters $\mat{V}^{(k)}$ for $k = 1, 2, \dots, M$ in the maximization problem \eqref{eq:CCA_max_problem_rplc_ast} can be obtained by solving the following maximization problem for each mode:
    }
    \begin{align}
        \label{eq:MCCA_max_problem_reformulated}
    &\max_{\substack{ \mat{V}^{(k)} \\ k=1,2,\dots, M }} \sum_{g=1}^G  \prod_{k=1}^{M} \tr\left\{  \trans{{\mat{V}^{(k)}}}  \mat{S}_{(g)}^{(k)} \mat{V}^{(k)} \trans{{\mat{V}^{(k)}}} \mat{S}_{(g)}^{(k)} \mat{V}^{(k)} \right\},\qq{s.t.} \trans{{\mat{V}^{(k)}}} \mat{V}^{(k)} = \mat{I}_{R_k}.
    \end{align}
\end{lemma}
However, we cannot simultaneously solve this problem for $\mat{V}^{(k)}, k=1,2, \dots, M$.
Thus, by summarizing the terms unrelated to $\mat{V}^{(k)}$ in the maximization problem (\ref{eq:MCCA_max_problem_reformulated}), we can obtain the maximization problem for $k$-th mode:
\begin{align}
    \label{eq:MCCA_max_problem_each_mode}
    \max_{\mat{V}^{(k)}} f_k({\mat{V}^{(k)}}) = \max_{\mat{V}^{(k)}}   \tr\left\{ \trans{{\mat{V}^{(k)}}} \mat{M}(\mat{V}^{(k)})  \mat{V}^{(k)} \right\},\qq{s.t.} \trans{{\mat{V}^{(k)}}} \mat{V}^{(k)} = \mat{I}_{R_k},
\end{align}
where $\mat{M}(\mat{V}^{(k)}) = \sum_{g=1}^G w_{(g)}^{(-k)} \mat{S}_{(g)}^{(k)} \mat{V}^{(k)} \trans{{\mat{V}^{(k)}}} \mat{S}_{(g)}^{(k)}$, in which $w_{(g)}^{(-k)}$ is given by
\begin{align}
    w_{(g)}^{(-k)} = \prod_{j \neq k} \tr\left\{  \trans{{\mat{V}^{(j)}}}  \mat{S}_{(g)}^{(j)} \mat{V}^{(j)} \trans{{\mat{V}^{(j)}}} \mat{S}_{(g)}^{(j)} \mat{V}^{(j)} \right\}.
\end{align}
Although an estimate of $\mat{V}^{(k)}$ can be obtained by solving the maximization problem (\ref{eq:MCCA_max_problem_each_mode}), this problem is non-convex, since $\mat{V}^{(k)}$ is assumed to be an orthogonal matrix. 
Thus, the maximization problem has several local maxima.
However, by choosing the initial values of parameters in the estimation near the global optimal solution, we can obtain the global optimal solution.
In Section \ref{sec:estimation}, we develop not only an estimation algorithm but also an initialization method for choosing the initial values of the parameters near the global optimal solution. 
The initialization method helps guarantee the convergence of our algorithm to the mode-wise global optimal solution.

\section{Estimation}
\label{sec:estimation}
Our estimation algorithm consists of two steps: initializing the parameters and iteratively updating the parameters.
The initialization step gives us the initial values of the parameters near the global optimal solution for each mode.
Next, by iteratively updating the parameters, we can monotonically increase the value of the objective function (\ref{eq:MCCA_max_problem_each_mode}) until convergence.

\subsection{Initialization}
\label{sec:initialization}
The first step is to initialize the parameters $\mat{V}^{(k)}$ for each mode.
We define an objective function $f_k'({\mat{V}^{(k)}}) = \tr\left\{ \trans{{\mat{V}^{(k)}}} \mat{M}\left(\mat{I}^{(k)}\right) \mat{V}^{(k)} \right\}$ for $k=1,\ldots,M$, where $\mat{M}\left(\mat{I}^{(k)}\right) = \sum_{g=1}^G w_{(g)}^{(-k)} \mat{S}_{(g)}^{(k)} \mat{S}_{(g)}^{(k)}$. 
Next, we  adopt a maximizer of $f_k'({\mat{V}^{(k)}})$ as initial values of the parameters ${\mat{V}^{(k)}}$. 
To obtain the maximizer, we need an initial value of $\bm{w}^{(k)} = \qty[w_{(1)}^{(-k)}, w_{(2)}^{(-k)}, \dots, w_{(G)}^{(-k)}]$. The initial value for $\bm{w}^{(k)}$ is obtained by solving the quadratic programming problem
\begin{align}
    \label{eq:QP_problem}
    \min_{\bm{w}^{(k)}} {\bm{w}^{(k)}}^\top \bm{\lambda}_0^{(k)} {\bm{\lambda}_0^{(k)}}^\top \bm{w}^{(k)}, \qq{s.t.} \bm{w}^{(k)} > \bm{0},\ {\bm{w}^{(k)}}^\top \bm{\lambda}_1^{(k)} {\bm{\lambda}_1^{(k)}}^\top \bm{w}^{(k)} = 1,
\end{align}
where 
\begin{align}
    \label{eq:lambda_0_1}
    \bm{\lambda}_0^{(k)} &= \qty[ \sum_{i=R_k + 1}^{P_k} {\lambda_{(1)i}^{(k)}}, \sum_{i=R_k + 1}^{P_k} {\lambda_{(2)i}^{(k)}}, \dots, \sum_{i=R_k + 1}^{P_k} {\lambda_{(G)i}^{(k)}} ]^\top, \nonumber \\
    \bm{\lambda}_1^{(k)} &= \qty[ \sum_{i=1}^{P_k} {\lambda_{(1)i}^{(k)}}, \sum_{i=1}^{P_k} {\lambda_{(2)i}^{(k)}}, \dots, \sum_{i=1}^{P_k} {\lambda_{(G)i}^{(k)}} ]^\top,
\end{align}
in which $\lambda_{(g)i}^{(j)}$ is the $i$-th largest eigenvalue of $\mat{S}_{(g)}^{(j)}\mat{S}_{(g)}^{(j)}$.

Using the initial value of $\bm{w}^{(k)}$, we can obtain the initial value of the parameter $\mat{V}_0^{(k)}$ by maximizing  $f_k'({\mat{V}^{(k)}})$ for each mode. 
The maximizer consists of $R_k$ eigenvectors, corresponding to the $R_k$ largest eigenvalues, obtained by eigenvalue decomposition of $\mat{M}\left(\mat{I}^{(k)}\right)$.
The theoretical justification for this initialization will be discussed in Section \ref{sec:theory}.

\subsection{Iterative Update of Parameters}
\label{sec:IterativeUpdateofParameters}
The second step is to update parameters $\mat{V}^{(k)}$ for each mode.
We update parameters such that the objective function $f_k (\mat{V}^{(k)})$ is maximized.
Let $\mat{V}_s^{(k)}$ be the value of $\mat{V}^{(k)}$ at step $s$. 
Then, we solve the surrogate maximization problem
\begin{align}
    \label{eq:updating_problem}
    \max_{\mat{V}_{s+1}^{(k)}}   \tr\left\{ \trans{{\mat{V}_{s+1}^{(k)}}} \mat{M}(\mat{V}_{s}^{(k)})  \mat{V}_{s+1}^{(k)} \right\},\qq{s.t.} \trans{{\mat{V}_{s+1}^{(k)}}} \mat{V}_{s+1}^{(k)} = \mat{I}_{R_k}.
\end{align}
The solution of (\ref{eq:updating_problem}) consists of $R_k$ eigenvectors, corresponding to the $R_k$ largest eigenvalues, obtained by eigenvalue decomposition of $\mat{M}(\mat{V}_{s}^{(k)})$.
By iteratively updating the parameters, the objective function $f_k (\mat{V}^{(k)})$ is monotonically increased, which allows it to be maximized.
The monotonically increasing property will be discussed in Section \ref{sec:theory}.

Our estimation procedure comprises the above estimation steps.
The procedure is summarized as Algorithm \ref{alg:algorithm_MCCA}.
\begin{algorithm}[H]
    \caption{Iteratively updating algorithm via eigenvalue decomposition}
    \label{alg:algorithm_MCCA}
    \begin{algorithmic}[1]
    \Require $M$-th order tensor dataset $\left\{ \mathcal{X}_{(g)} \in \mathbb{R}^{P_1 \times P_2 \times \dots \times P_M \times N_g},g=1,2,\dots,G \right\}$.
        \State \textbf{Calculate covariance matrix for tensors: }$\mat{S}_{(g)}^*$ via (\ref{eq:covariance_matrix}) and (\ref{eq:covariance_matrix_each_mode}).

        \State \textbf{Step 1 Initialization: }
        \State $\bm{w}^{(k)} \gets$ the solution of quadratic programming problem (\ref{eq:QP_problem}), $k = 1,2,\dots, M$.
        \State $\mat{V}_0^{(k)} \gets$ $R_k$ eigenvectors obtained by the eigenvalue decomposition of $\mat{M}\left(\mat{I}^{(k)}\right)$,\quad $k=1,2,\dots,M$.

        \State $\mat{\Lambda}_{(g)}^{(k)} \gets \trans{{\mat{V}^{(k)}}} \mat{S}_{(g)}^{(k)} \mat{V}^{(k)},\quad k=1,2,\dots,M ; g=1,2,\dots,G$.

        \State \textbf{Step 2 Updating parameters: }

        \For {$s = 1, 2,\dots$}

        \State \textbf{Update $\mat{V}^{(k)}$:} $\mat{V}_{s+1}^{(k)} \gets$ $R_k$ eigenvectors obtained by eigenvalue decomposition of $\mat{M}\qty(\mat{V}_s^{(k)})$,\quad $k=1,2,\dots,M$.

        \State \textbf{Update $\mat{\Lambda}_{(g)}^{(k)}$:} $\mat{\Lambda}_{(g)}^{(k)} \gets \trans{{\mat{V}_{s+1}^{(k)}}} \mat{S}_{(g)}^{(k)} \mat{V}_{s+1}^{(k)},\quad k=1,2,\dots,M ;  g=1,2,\dots,G$.

        \EndFor
        \State \textbf{return} $\mat{V}^{(k)} \in \mathbb{R}^{P_k \times R_k}, \mat{\Lambda}_{(g)}^{(k)} \in \mathbb{S}_+^{R_k},\quad k=1,2,\dots,M ;  g=1,2,\dots,G$.
    \end{algorithmic}
\end{algorithm}

\section{Theory}
\label{sec:theory}
This section presents the theoretical and computational analyses for Algorithm \ref{alg:algorithm_MCCA}.
Theoretical analyses consist of two steps.
First, we prove that the initial values of parameters obtained in Section \ref{sec:initialization} are relatively close to the global optimal solution.
If the initial values are close to the global maximum, then we can obtain the global optimal solution even if the maximization problem is non-convex.
Second, we prove that the iterative updates of the parameters in Section \ref{sec:IterativeUpdateofParameters} monotonically increase the value of objective function (\ref{eq:MCCA_max_problem_each_mode}) by solving the surrogate problem (\ref{eq:updating_problem}).
From the monotonically increasing property, the estimated parameters always converge at a stationary point.
The combination of these two results enables us to obtain the mode-wise global optimal solution.
In the computational analysis, we calculate computational cost for MCCA and then compare the cost with conventional methods.
By comparing the costs, we investigate the computational efficacy of MCCA.

\subsection{Analysis of Upper and Lower Bounds}
\label{subsec:analysis_of_upper_and_lower_bounds}
The aim of this subsection is to provide the upper and lower bounds of the maximization problem (\ref{eq:MCCA_max_problem_each_mode}).
From the bounds, we find that the initial values in Section 4.1 are relatively close to the global optimal solution.
Before providing the bounds, we define a contraction ratio. 
\begin{definition}
    \textit{
    Let $f_k'^{\max}$ be the global maximum of $f_k'(\mat{V}^{(k)})$ and $M^{(k)} = \tr \qty{\mat{M}\qty(\mat{I}^{(k)}) }$. 
    Then a contraction ratio of data for $k$-th mode is defined by
    \begin{align}
    \label{def:contraction_ratio}
        \alpha^{(k)} =  \frac{f_k'^{\max}}{M^{(k)}} =  \frac{ \tr\left\{ \trans{{\mat{V}_0^{(k)}}} \mat{M}\left(\mat{I}^{(k)}\right) \mat{V}_0^{(k)} \right\}}{\tr \left\{\mat{M}\left(\mat{I}^{(k)}\right)\right\}}.
    \end{align}
}
\end{definition}
Note that a contraction ratio $\alpha^{(k)}$ satisfies $0 \leq  \alpha^{(k)} \leq 1$ and $\alpha^{(k)} = 1$ if and only if $R_k = P_k$.

Using $f_k'^{\max}$ and the contraction ratio $\alpha^{(k)}$, we have the following theorem that reveals the upper and lower bounds of the global maximum in the problem (\ref{eq:MCCA_max_problem_each_mode}).
\begin{theorem}
    \label{theorem:upper_and_lower_bound_max}
    \textit{
    Let $f_k^{\max}$ be the global maximum of $f_k(\mat{V}^{(k)})$. Then 
    \begin{align}
        \alpha^{(k)} f_k'^{\max} \leq f_k^{\max} \leq f_k'^{\max},
    \end{align}
    where $\alpha^{(k)}$ is the contraction ratio defined in (\ref{def:contraction_ratio}) and $f_k'^{\max}$ is the global maximum of $f_k'(\mat{V}^{(k)})$.
    }
\end{theorem}
This theorem indicates that $f_k'^{\max} \to f_k^{\max}$ when $\alpha^{(k)} \to 1$. 
Thus, it is important to obtain an $\alpha^{(k)}$ that is as close as possible to one. 
Since $\alpha^{(k)}$ depends on $\mat{V}_0^{(k)}$ and ${\bm w}^{(k)}$, $\mat{V}_0^{(k)}$ depends on ${\bm w}^{(k)}$. 
From this dependency, if we could set the initial value of ${\bm w}^{(k)}$ such that $\alpha^{(k)}$ is as large as possible, then we could obtain an initial value of $\mat{V}_0^{(k)}$ that attains a value near $f_k^{\max}$. 
The following theorem shows that we can compute the initial value of ${\bm w}^{(k)}$ such that $\alpha^{(k)}$ is maximized. 
\begin{theorem}
    \label{theorem:optimal_alpha}
    \textit{
    Let $\bm{\lambda}_0^{(k)}$ and $\bm{\lambda}_1^{(k)}$ be the vectors consisting of eigenvalues defined in (\ref{eq:lambda_0_1}). For $\bm{w}^{(k)} = \qty[w_{(1)}^{(-k)}, w_{(2)}^{(-k)}, \dots, w_{(G)}^{(-k)}] \ (k=1,2,\dots,M)$, suppose that the estimate $\hat{\bm{w}}^{(k)}$ is obtained by solving (\ref{eq:QP_problem}) for $k=1,2,\dots,M$. Then $\hat{\bm{w}}^{(k)}$  maximizes $\alpha^{(k)}$.}
\end{theorem}
In fact, $\alpha^{(k)}$ is very close to one with the initial values given in Theorem \ref{theorem:optimal_alpha} even if $R_k$ is small.
This resembles the cumulative contribution ratio in PCA. 


\subsection{Convergence Analysis}
We next verify that our proposed procedure for iteratively updating parameters maximizes the optimization problem (\ref{eq:MCCA_max_problem_each_mode}).
In Algorithm \ref{alg:algorithm_MCCA}, the parameter $\mat{V}_{s+1}^{(k)}$ can be obtained by solving the surrogate maximization problem (\ref{eq:updating_problem}).
The following Theorem \ref{theorem:update_guarantee_increase} shows that we can monotonically increase the value of the function $f_k (\mat{V}^{(k)})$ in \eqref{eq:MCCA_max_problem_each_mode} by Algorithm \ref{alg:algorithm_MCCA}.
\begin{theorem}
    \label{theorem:update_guarantee_increase}
    \textit{
    Let $\mat{V}_{s+1}^{(k)} $ be $R_k$ eigenvectors, corresponding to the  $R_k$ largest eigenvalues, obtained by eigenvalue decomposition of $\mat{M}(\mat{V}_s^{(k)})$. 
    Then 
    \begin{align}
        f_k(\mat{V}_s^{(k)}) \leq f_k(\mat{V}_{s+1}^{(k)}).
    \end{align}
    }
\end{theorem}
From Theorem \ref{theorem:upper_and_lower_bound_max}, we obtain initial values of the parameters that are near the global optimal solution.
By combining Theorem \ref{theorem:upper_and_lower_bound_max} and Theorem \ref{theorem:update_guarantee_increase}, the solution from Algorithm \ref{alg:algorithm_MCCA} can be characterized by the following corollary.
\begin{corollary}
    \textit{
    Consider the maximization problem (\ref{eq:MCCA_max_problem_each_mode}).
    Suppose that the initial value of the parameter is obtained by  $\mat{V}_0^{(k)}$ = $\argmax_{\mat{V}^{(k)}} \widetilde{f_k}' (\mat{V}^{(k)})$ and the parameter $\mat{V}_s^{(k)}$ is repeatedly updated by Algorithm \ref{alg:algorithm_MCCA}.
    Then the mode-wise global maximum for the maximization problem (\ref{eq:MCCA_max_problem_each_mode}) is achieved when all the contraction ratios $\alpha^{(k)}$ for $k = 1, 2, \dots, M$ go to one.}
\end{corollary}
Algorithm \ref{alg:algorithm_MCCA} does not guarantee the global solution, due to the fundamental problem of non-convexity, but it is enough for pragmatic purposes.
We will investigate the issue of convergence to global solution through numerical studies in Section \ref{sec:BehaviorofContractionRatio}. 

\subsection{Computational Analysis}
First, we analyze the computational cost.
To simplify the analysis, we assume $P = \argmax_j P_j$ for $j=1,2,\dots,M$.
This implies that $P$ is the upper bound of $R_j$ for all $j$.
We then calculate the upper bound of the computational complexity.

The expensive computations of the each iteration in Algorithm \ref{alg:algorithm_MCCA} consist of three parts: the formulation of $\mat{M}(\mat{V}_s^{(k)})$, the eigenvalue decomposition of $\mat{M}(\mat{V}_s^{(k)})$, and updating latent covariance matrices $\mat{\Lambda}_g^{(k)}$.
These steps are $O(GM^2P^3)$, $O(P^3)$, and $O(GMP^3)$, respectively.
The total computational complexity per iteration is then  $O(GM^2P^3)$.
This indicates that the MCCA algorithm is not limited by the sample size.
In contrast, the MPCA algorithm is affecred by the sample size \citep{Lu_2008}.

Next, we analyze the memory requirement of Algorithm \ref{alg:algorithm_MCCA}.
MCCA represents the original tensor data with fewer parameters by projecting the data onto a lower-dimensional space.
This requires the $P_k \times R_k$ projection matrices $\mat{V}^{(k)}$ for $k=1,2,\dots,M$.
MCCA projects the data a the size of $N  \qty( \prod_{k=1}^M P_k)$ to $N  \qty(\prod_{k=1}^M R_k )$, where $N = \sum_{g=1}^G N_g$.
Thus, the required size for the parameters is $\sum_{k=1}^{M} P_k R_k + N \qty( \prod_{k=1}^M R_k )$.
MPCA requires the same amount of memory as MCCA.
Meanwhile, CCA and PCA need a projection matrix, which is size $R \qty( \prod_{k=1}^{M} P_k )$.
The required size for the parameters is then $R \qty( \prod_{k=1}^{M} P_k ) + NR$.
It should be noted that MCCA and MPCA require a large amount of memory when the number of modes in a dataset is large, but their memory requirements are much smaller than those of CCA and PCA.

\section{Experiment}
\label{sec:Experiment}
To demonstrate the efficacy of MCCA, we applied MCCA, PCA, CCA, and MPCA to image compression tasks.

\subsection{Experimental Setting}
For the experiments, we prepared the following three image datasets:
\begin{description}
    \item[MNIST dataset] consists of data of hand written digits $0, 1, \dots, 9$ at image sizes of $28 \times 28$ pixels.
    The dataset includes a training dataset of 60,000 images and a test dataset of 10,000 images.
    We used the first 10 training images of the dataset for each group.
    The MNIST dataset \citep{Lecun_1998} is available at \url{http://yann.lecun.com/exdb/mnist/}.
    \item[AT\&T (ORL) face dataset] contains gray-scale facial images of 40 people.
    The dataset has 10 images sized $92 \times 112$ pixels for each person.
    We used images resized by a factor of 0.5 in order to improve the efficiency of the experiment. 
    The AT\&T face dataset is available at \url{https://git-disl.github.io/GTDLBench/datasets/att_face_dataset/}.
    \item[Cropped AR database] has color facial images of 100 people.
    These images are cropped around the face. The size of images is $120 \times 165 \times 3$ pixels.
    The dataset contains 26 images in each group, 12 of which are images of people wearing sunglasses or scarves.
    We used the cropped facial images of 50 males which were not wearing sunglasses or scarves.
    Due to memory limitations, we resized these images by a factor of 0.25.
    The AR database \citep{Martinez_1998, Martinez_2001} is available at \url{http://www2.ece.ohio-state.edu/~aleix/ARdatabase.html}.
\end{description}
The dataset characteristics are summarized in Table \ref{tb:dataset_all}.
\begin{table}[H]
\caption{Summary of the datasets.}
\label{tb:dataset_all}
\begin{tabular}{lcccc}
\hline
Dataset & Group size & Sample size (/group) & Number of dimensions  & Number of groups \\
\hline
MNIST  & Small & 10 & $28 \times 28 =  784$ & 10 \\
\hline
\multirow{3}{*}{AT\&T(ORL)} & Small & \multirow{3}{*}{10} & \multirow{3}{*}{$46 \times 56 =  2576$} & 10\\
 & Medium & &  & 20 \\
 & Large & &  & 40 \\
\hline
\multirow{3}{*}{Cropped AR} & Small &\multirow{3}{*}{14} & \multirow{3}{*}{$ 30 \times 41 \times 3 = 7380
$} & 10\\
 & Medium & &  & 25  \\
 & Large & &  & 50\\
 \hline
\end{tabular}
\end{table}

To compress these images, we performed dimensionality reductions by MCCA, PCA, CCA, and MPCA, as follows.
We vectorized the tensor dataset before performing PCA and CCA.
In MCCA, the images were compressed and reconstructed according to the following steps.
\begin{enumerate}
    \item Prepare the multiple image datasets  $\mathcal{X}_{(g)} \in \mathbb{R}^{P_1 \times P_2 \times \dots \times P_M \times N_g}$ for $g=1, 2, \dots, G$.
    \item Compute the covariance matrix of $\mathcal{X}_{(g)} $ for $g=1,2,\dots,G$.
    \item From these covariance matrices, compute the linear transformation matrices  $\mat{V}_i \in \mathbb{R}^{P_i \times R_i}$ for $i =1, 2,\dots, M$ for mapping to the $(R_1, R_2, \dots, R_M)$-dimensional latent space.
    \item Map the $i$-th sample $\mathcal{X}_{(g)i}$ to $\mathcal{X}_{(g)i} \times_1 \mat{V}_1 \times_2 \mat{V}_2 \cdots \times_M \mat{V}_M \in \mathbb{R}^{R_1 \times R_2 \times \dots \times R_M}$, where the operator $\times_i$ is the $i$-mode product of tensor \citep{Kolda_2009}.
    \item Reconstruct $i$-th sample $\tilde{\mathcal{X}}_{(g)i}$ = $\mathcal{X}_{(g)i} \times_1 \mat{V}_1 \mat{V}_1^\top \times_2 \mat{V}_2 \mat{V}_2^\top \cdots \times_M \mat{V}_M \mat{V}_M^\top$.  
\end{enumerate}
Meanwhile, PCA and MPCA each require a single dataset.
Thus, we aggregated the datasets as $\mathcal{X} = [\mathcal{X}_{(1)}, \mathcal{X}_{(2)}, \dots, \mathcal{X}_{(G)}] \in \mathbb{R}^{P_1 \times P_2 \times \dots \times P_M \times \sum_{g=1}^G Ng}$ and performed PCA and MPCA for the dataset $\mathcal{X}$. 

\subsection{Performance Assessment}
For MCCA and MPCA, the reduced dimensions $R_1$ and $R_2$ were chosen as the same number, and then we fixed $R_3$ as two. 
All computations were performed by the software \texttt{R} (ver. 3.6) \citep{R_core_team}.
In the initialization of MCCA, solving the quadratic programming problem was carried out using the function \texttt{ipop} in the package \texttt{kernlab}.
MPCA was implemented as the function \texttt{mpca} in the package \texttt{rTensor}. 
The implementations of MCCA, PCA, and CCA are available at \url{https://github.com/yoshikawa-kohei/MCCA}.

To assess their performances, we calculated the reconstruction error rate (RER) under the same compression ratio (CR).
RER is defined by
\begin{align}
    \label{def:RER}
    \mathrm{RER} = \frac{ \norm{\mathcal{X} - \widetilde{\mathcal{X}}}_F^2 }{\norm{\mathcal{X}}_F^2},
\end{align}
where $\widetilde{\mathcal{X}} = [\widetilde{\mathcal{X}}_{(1)},\widetilde{\mathcal{X}}_{(2)}, \dots, \widetilde{\mathcal{X}}_{(G)}]$ is the aggregated dataset of reconstructed tensors  $\widetilde{\mathcal{X}}_{(g)} = [\tilde{\mathcal{X}}_{(g)1},\tilde{\mathcal{X}}_{(g)2},\dots, \tilde{\mathcal{X}}_{(g)N_g}]$ for $g=1,2, \dots, G$ and $\| \mathcal{X} \|_F$ is the norm of a tensor $\mathcal{X} \in \mathbb{R}^{P_1 \times P_2 \times \dots \times P_M}$ computed by
\begin{align}
    \| \mathcal{X} \|_F = \sqrt{ \sum_{p_1 = 1}^{P_1} \sum_{p_2 = 1}^{P_2} \cdots \sum_{p_M = 1}^{P_M} x_{p_1, p_2, \dots, p_M}^2},
\end{align}
in which $x_{p_1, p_2, \dots, p_M}$ is an element $(p_1, p_2, \dots, p_M)$ of $\mathcal{X}$.
In addition, we defined CR as
\begin{align}
    \label{eq:comp_ratio}
    \mathrm{CR} = \frac{\text{\# \{The number of required parameters\}}}{N \cdot \prod_{k=1}^M P_k}.
\end{align}
The number of parameters required for MCCA and MPCA is $\sum_{k=1}^{M} P_k R_k  + N \qty( \prod_{k=1}^M R_k )$, whereas that for CCA and PCA is $R \qty( \prod_{k=1}^{M} P_k) + NR$.

\begin{figure}[H]
\centering
    \subcaptionbox{\label{fig:RER_ORL_small}}{
        \includegraphics[width=0.45\linewidth]{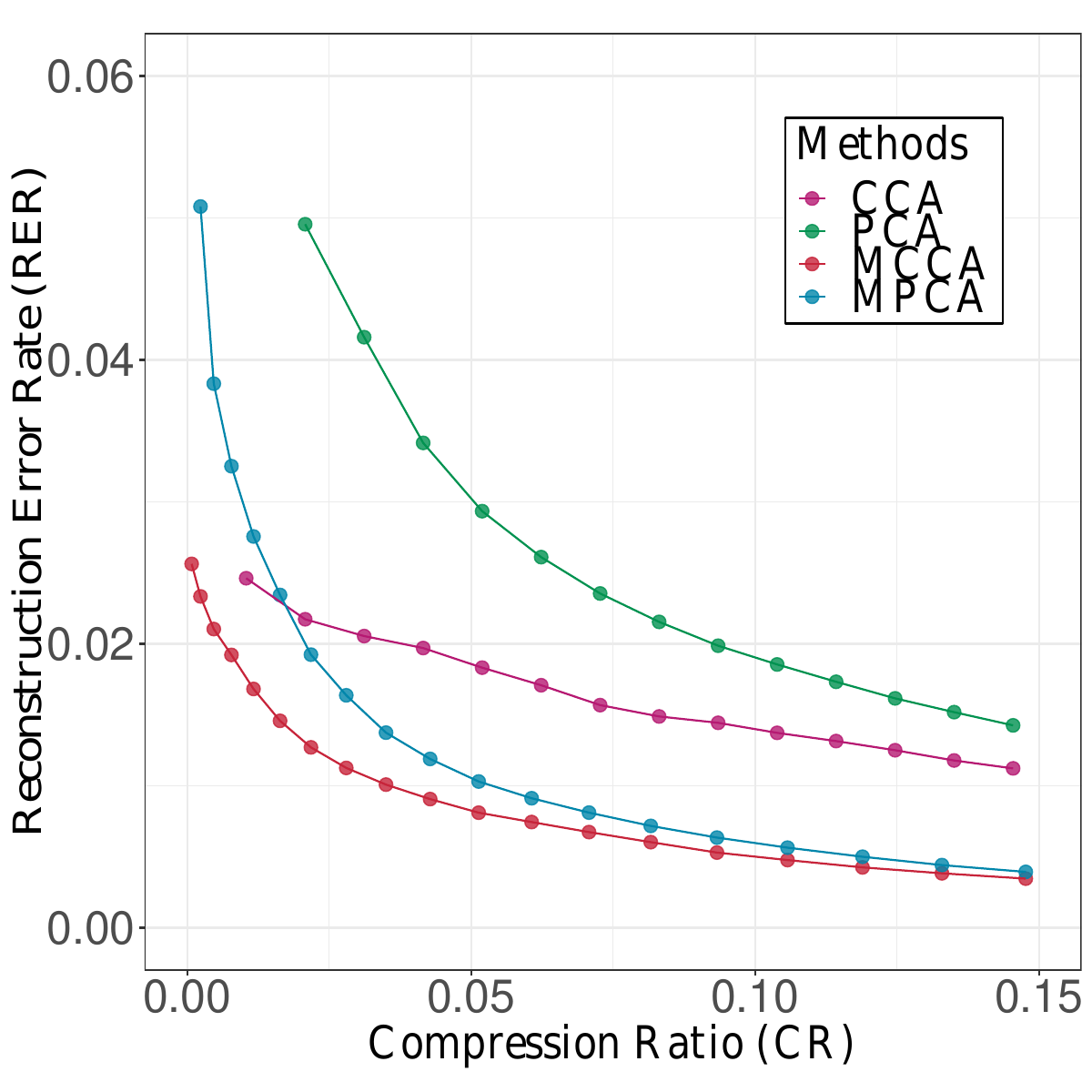}
    }
    \subcaptionbox{\label{fig:RER_ORL_medium}}{
        \includegraphics[width=0.45\linewidth]{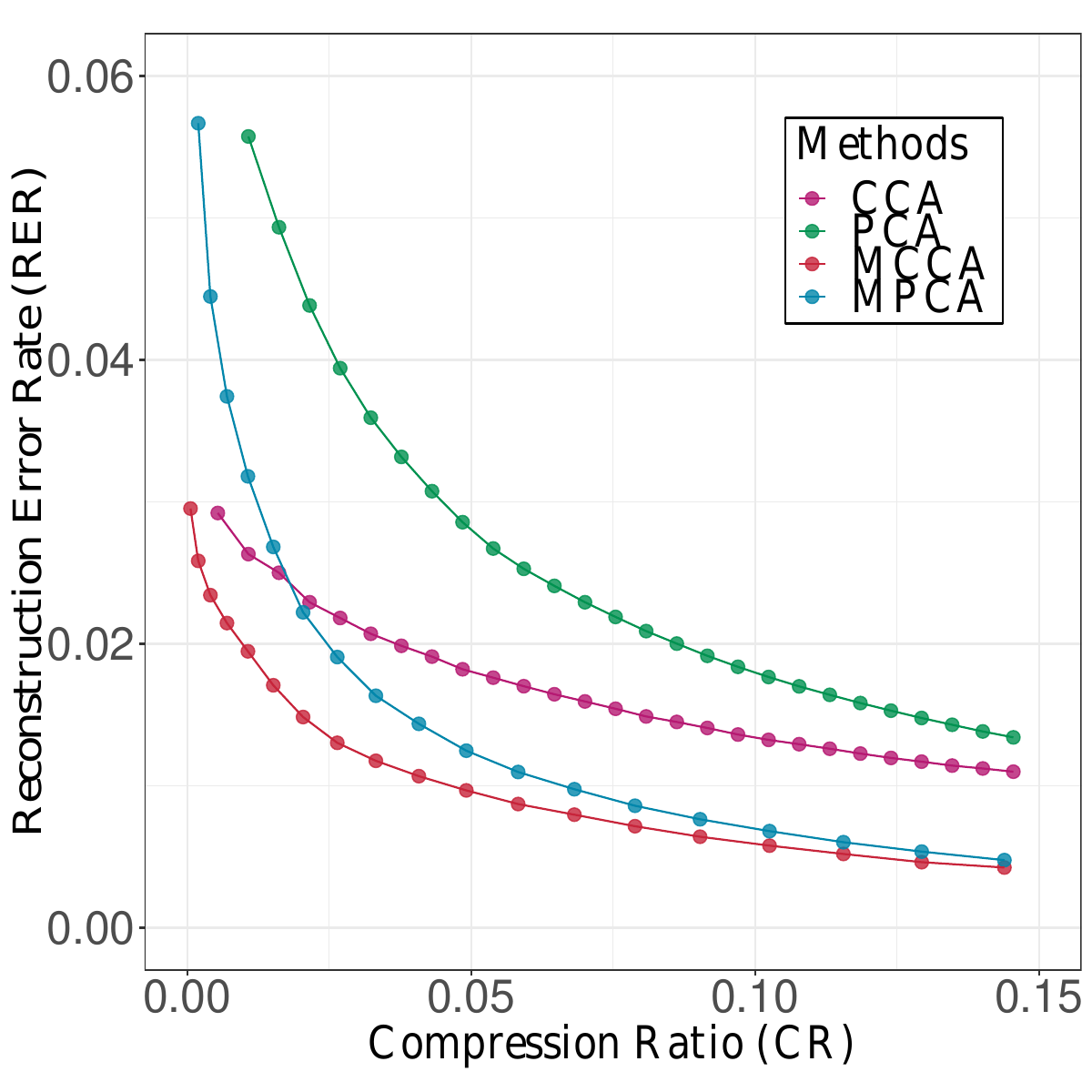}
    }
    \subcaptionbox{\label{fig:RER_ORL_large}}{
        \includegraphics[width=0.45\linewidth]{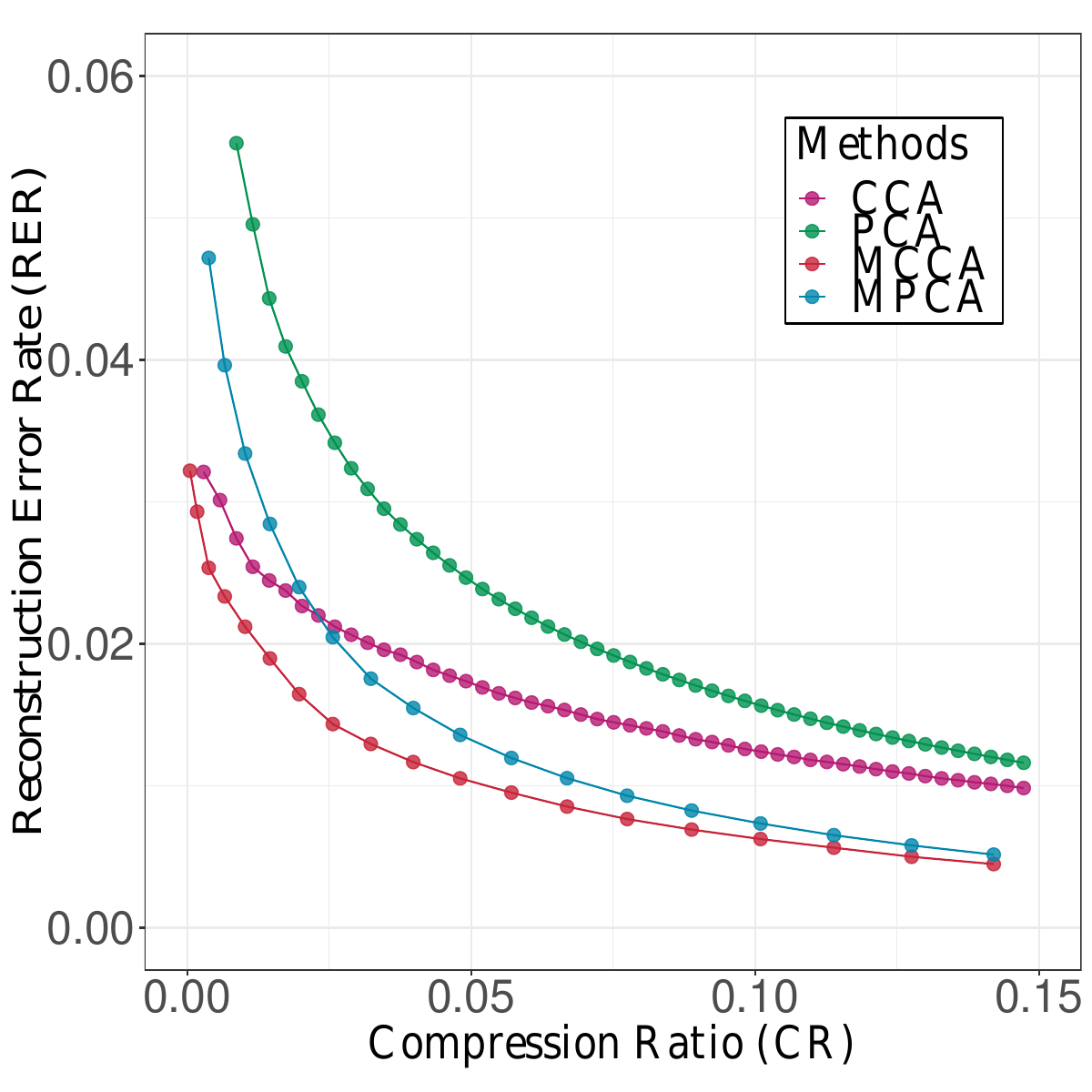}
    }
    \caption{Plots of RER versus CR for the AT\&T(ORL) dataset of various group sizes: (a) Small; (b) Medium; and (c) Large.}
    \label{fig:ORL_all_results}
\end{figure}

Figure \ref{fig:ORL_all_results} plots RER obtained by estimating various reduced dimensions for the AT\&T(ORL) dataset with group sizes of small, medium, and large.
As the figures for the results of the other datasets were similar to Figure 1, we show them in the supplementary materials S1. 

From Figure \ref{fig:ORL_all_results}, we observe that the RER of MCCA is the smallest for any value of CR.
This indicates that the MCCA performs better than the other methods.
In addition, note that CCA performs better than MPCA only for fairly small values of CR, even though it is a method for vector data, whereas MPCA performs better for larger values of CR.
This implies the limitations of CCA for vector data.

Next we cobsider group size by comparing (a), (b), and (c) in Figure \ref{fig:ORL_all_results}. 
The value of CR at the intersection of CCA and MPCA increases with increasing the group size. 
This indicates that MPCA has more trouble extracting an appropriate latent space as the group size increases. 
Since MPCA does not consider the group structure, it is not possible to properly estimate the covariance structure when the group size is large.

\subsection{Behavior of Contraction Ratio}
\label{sec:BehaviorofContractionRatio}
We examined the behavior of contraction ratio $\alpha^{(k)}$. 
We performed MCCA on the AT\&T(ORL) dataset with the medium group size and computed $\alpha^{(1)}$ and $\alpha^{(2)}$ with the various pairs of reduced dimensions $(R_1, R_2) \in \{1,2,\dots, 25\} \times \{1,2,\dots, 25\}$.

\begin{figure}[H]
    \centering
    \includegraphics[width=\linewidth]{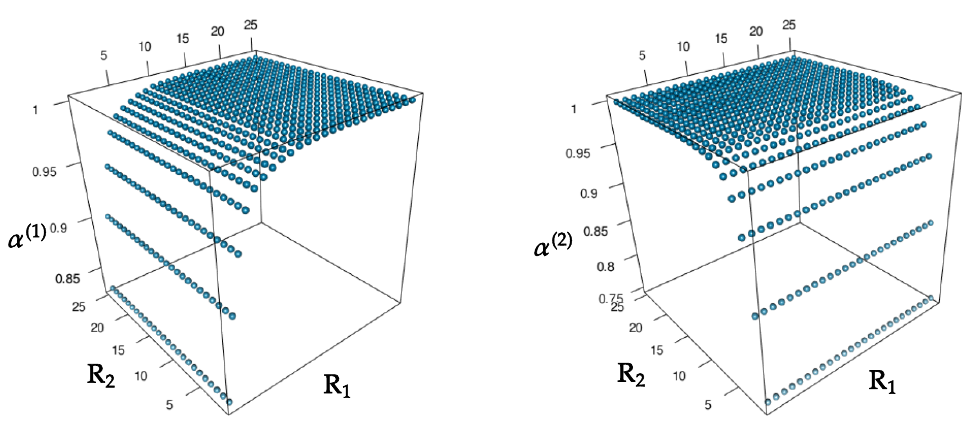}
    \caption{$\alpha^{(1)}$ and $\alpha^{(2)}$ versus pairs of reduced dimensions $(R_1, R_2)$.}
    \label{fig:alpha_3d}
\end{figure}

\begin{figure}[H]
    \centering
    \includegraphics[width=\linewidth]{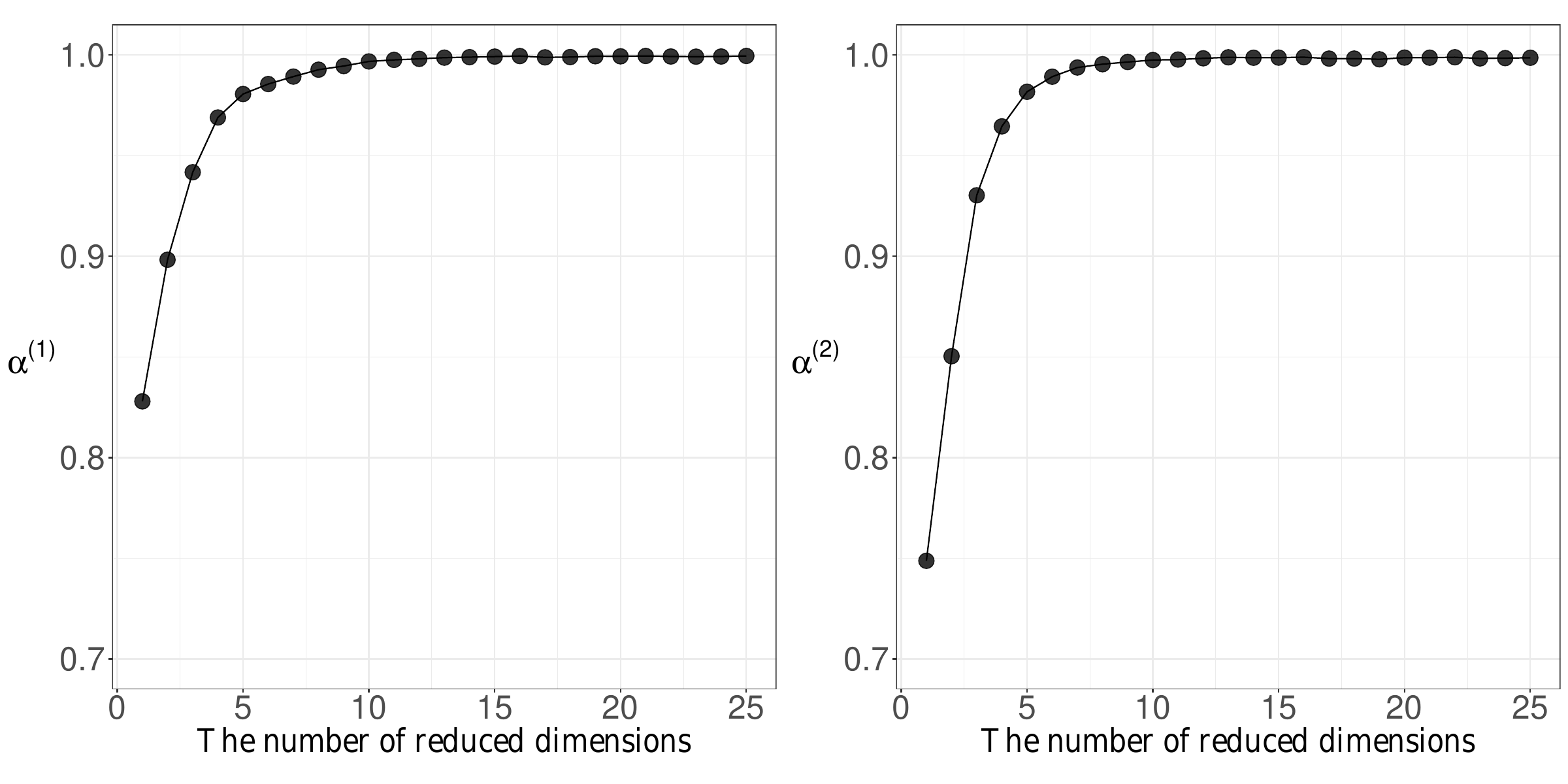}
    \caption{$\alpha^{(1)}$ and $\alpha^{(2)}$ versus $R_1$ and $R_2$, respectively.}
    \label{fig:alpha_2d}
\end{figure}

Figure \ref{fig:alpha_3d} shows the values of $\alpha^{(1)}$ and $\alpha^{(2)}$ for all pairs of $R_1$ and $R_2$. 
As shown, $\alpha^{(1)}$ and $\alpha^{(2)}$ were invariant to variations in $R_2$ and $R_1$, respectively.
Therefore, to facilitate visualization of changes in $\alpha^{(k)}$, Figure \ref{fig:alpha_2d} shows $\alpha^{(1)}$ and $\alpha^{(2)}$ for, respectively, $R_2=1$ and $R_1 = 1$.
From these, we observe that when both $R_1$ and $R_2$ are greater than 8, both $\alpha^{(1)}$ and $\alpha^{(2)}$ are close to one. 

\subsection{Efficacy of Solving the Quadratic Programming Problem}
We investigated the usefulness of determining the initial value of $\bm{w}^{(k)}$ by solving the quadratic programming problem (\ref{eq:QP_problem}). 
We applied MCCA to the AT\&T(ORL) dataset with the small, medium, and large number of groups. 
In addition, we also used the smaller group size of three.
For determining the initial value of $\bm{w}^{(k)}$, we consider three methods: solving the quadratic programming problem (\ref{eq:QP_problem}) (MCCA:QP), setting all values of $\bm{w}^{(k)}$ to one (MCCA:FIX), and setting the values by random sampling according to the uniform distribution $U(0,1)$ (MCCA:RANDOM).
We computed the $\alpha^{(k)}$ with the reduced dimensions $R_1 = R_2 \ (\in \qty{1,2,\dots, 10})$ for each of these methods.

To evaluate the performance of these methods, we compared the values of $\alpha^{(k)}$ and the number of iterations in the estimation.
The number of iterations in the estimation is the number of repetitions of lines 7 to 9 in Algorithm \ref{alg:algorithm_MCCA}.
For MCCA(RANDOM), we performed 50 trials and calculated averages of each of these indices.

\begin{figure}[H]
    \centering
    \includegraphics[width=\linewidth]{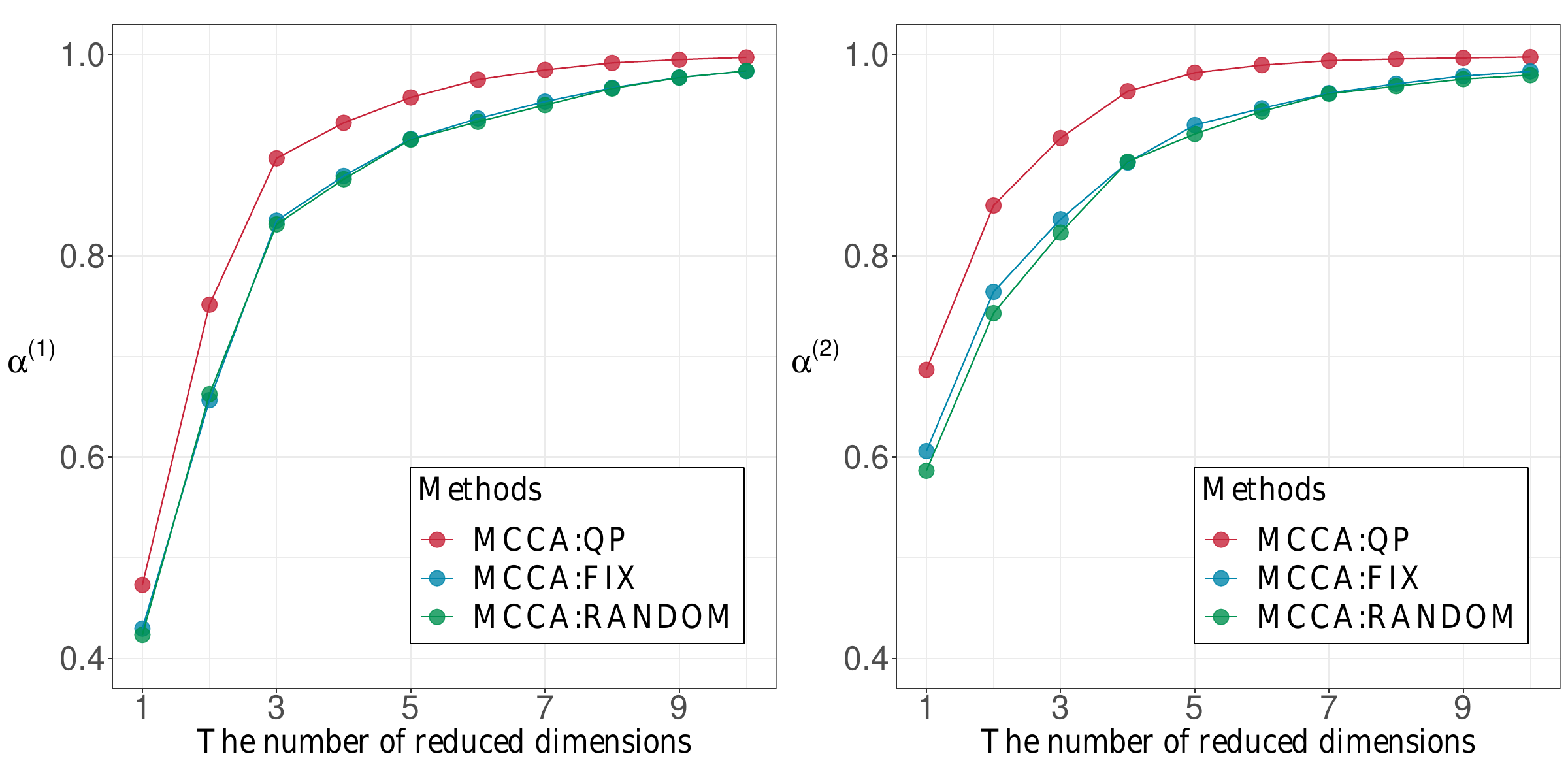}
    \caption{Comparisons of $\alpha^{(1)}$ and $\alpha^{(2)}$ computed by using the initial values obtained from the initializations MCCA:QP, MCCA:FIX, and MCCA:RANDOM with the AT\&T(ORL) dataset for a group size of 3.}
    \label{fig:qp_alpha_g3}
\end{figure}

\begin{figure}[H]
    \centering
    \includegraphics[width=\linewidth]{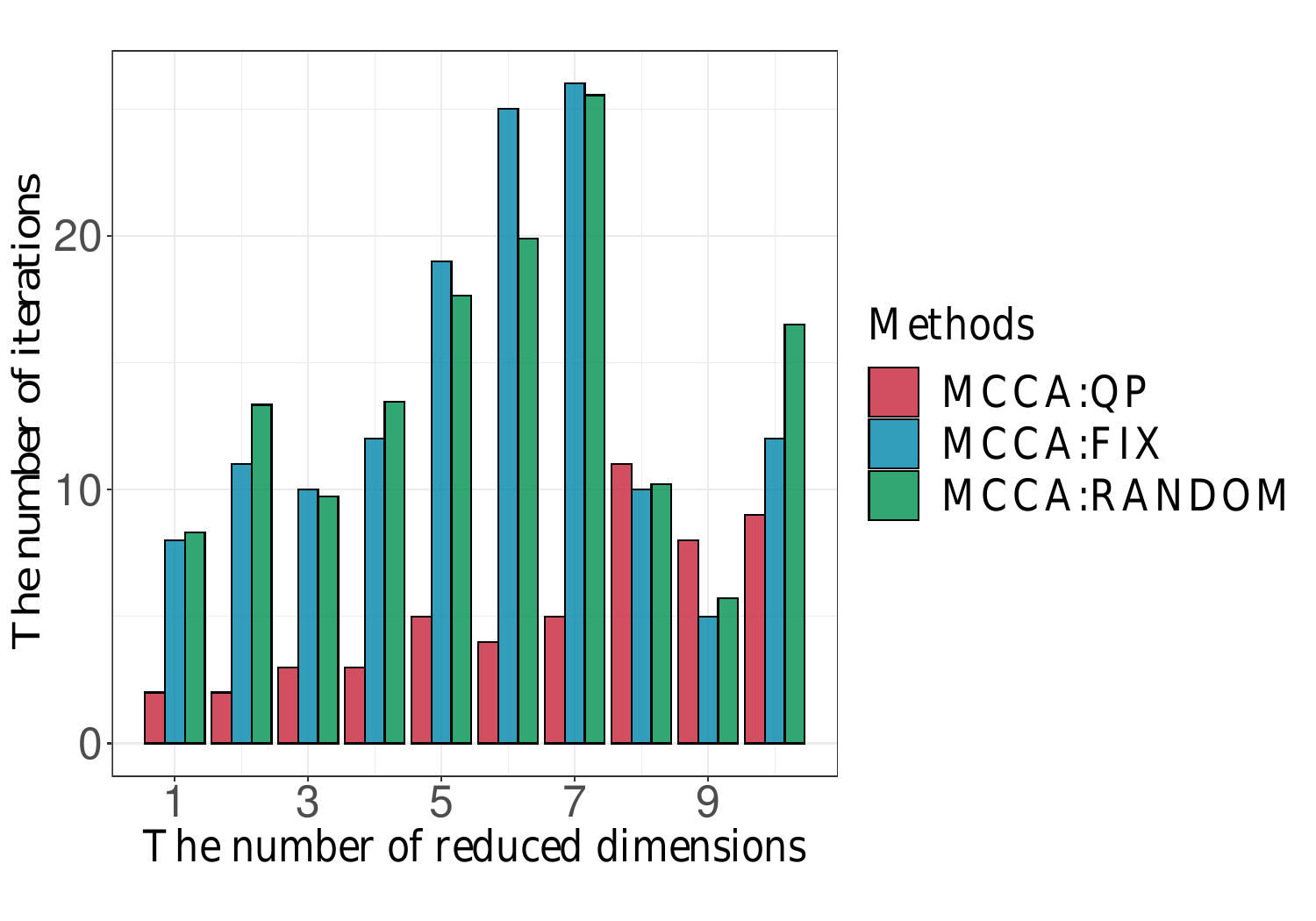}
    \caption{Comparison of the number of iterations when the initialization was performed by MCCA:QP, MCCA:FIX, and MCCA:RANDOM with the AT\&T(ORL) dataset for a group size of 3.}
    \label{fig:num_iter_g3}
\end{figure}

Figure \ref{fig:qp_alpha_g3} shows the comparisons of $\alpha^{(1)}$ and $\alpha^{(2)}$ when the initialization was performed by MCCA:QP, MCCA:FIX, and MCCA:RANDOM for AT\&T(ORL) dataset with a group size of 3.
It was confirmed that MCCA:QP provides the largest values of $\alpha^{(1)}$ and $\alpha^{(2)}$. 
Figure \ref{fig:num_iter_g3} shows that the number of iterations. 
MCCA:QP gives the smallest number of iterations for almost all values of the reduced dimensions.
This result indicates that MCCA:QP converges to a solution faster than the other initialization methods.
However, when the reduced dimension is greater than or equal to 8, the other methods are competitive with MCCA:QP.
A lack of difference in the number of iterations could result from the closeness of the initial values and the global optimal solution.
Note that when the $R_1$ and $R_2$ are greater than or equal to 8, $\alpha^{(1)}$ and $\alpha^{(2)}$ are sufficiently close to one, based on Figure \ref{fig:qp_alpha_g3}.
This indicates that the initial values are close to the global optimal solution obtained from Theorem \ref{theorem:upper_and_lower_bound_max}.
Hence, the result shows almost the same numbers of iterations for the three methods.

\begin{figure}[H]
    \centering
    \includegraphics[width=\linewidth]{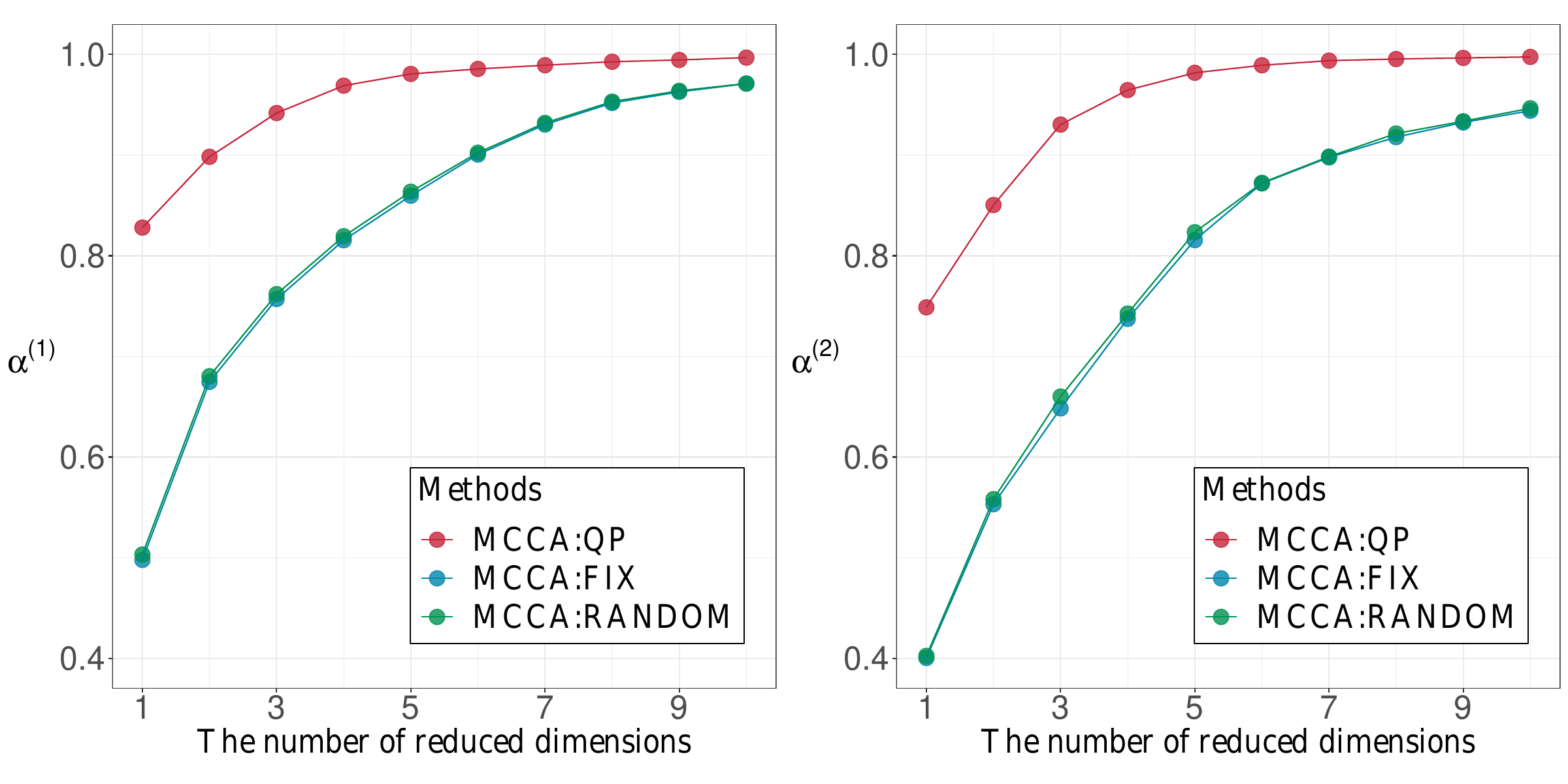}
    \caption{Comparisons of $\alpha^{(1)}$ and $\alpha^{(2)}$ computed using the initial values obtained from the initialization of MCCA:QP, MCCA:FIX, and MCCA:RANDOM with the AT\&T(ORL) dataset and the medium group size.}
    \label{fig:qp_alpha_g20}
\end{figure}

\begin{figure}[H]
    \centering
    \includegraphics[width=\linewidth]{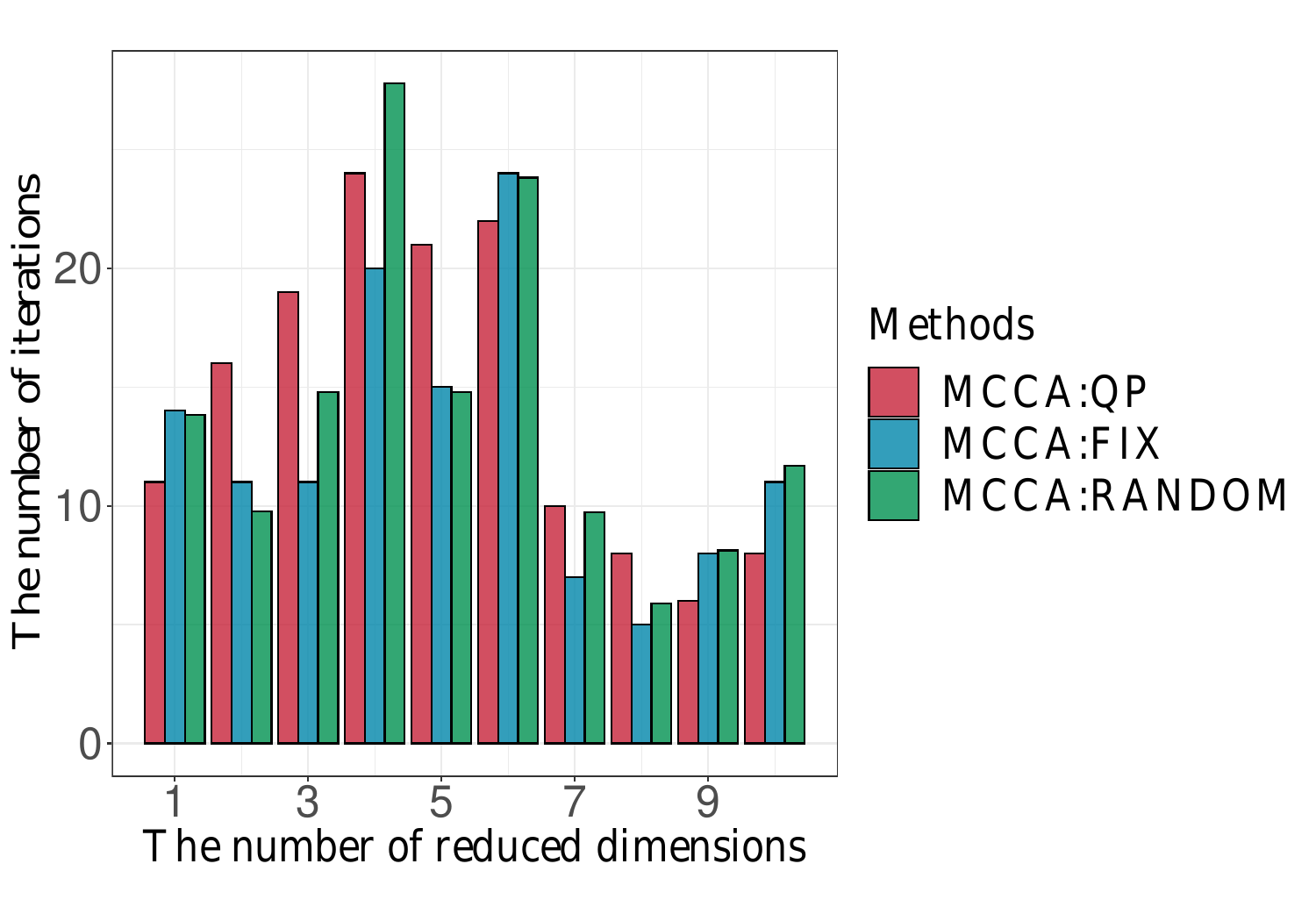}
    \caption{Comparison of the number of iterations when the initialization was perfomed by MCCA:QP, MCCA:FIX, and MCCA:RANDOM with the AT\&T(ORL) dataset and the medium group size.}
    \label{fig:num_iter_g20}
\end{figure}

Figures \ref{fig:qp_alpha_g20} and \ref{fig:num_iter_g20} show comparisons for the AT\&T(ORL) dataset with the medium group size.
Since the figures for the results of other group sizes are similar to Figures \ref{fig:qp_alpha_g20} and \ref{fig:num_iter_g20}, we show them in the supplementary mateirals S2.
Figure \ref{fig:qp_alpha_g20} shows results similar those in Figure \ref{fig:qp_alpha_g3}, whereas Figure \ref{fig:num_iter_g20} shows competitive performances for all reduced dimensions.

\section{Concluding Remarks}
We have developed the multilinear common components analysis (MCCA) by introducing a covariance structure based on the Kronecker product. 
To efficiently solve the non-convex optimization problem for MCCA, we have proposed an iteratively updating algorithm. 
The proposed algorithm exhibits some superior theoretical convergence properties.
Numerical experiments showed the usefulness of MCCA.

Specifically, MCCA was shown to be competitive among the initialization methods in terms of number of iterations.
As the number of groups increases, the overall number of samples increases. 
This may be why all methods required almost the same number of iterations for small, medium, and large number of groups. 
Note that, in this study, we used the Kronecker product representation to estimate the covariance matrix for tensor datasets. 
\citet{Greenewald_2019_JRSS} used the Kronecker sum representation for estimating the covariance matrix, and it would be interesting to extend the MCCA to this and other covariance representations.

\appendix
\section*{Appendices}
\section{Proof of Lemma \ref{lemma:MCCA_reformulate_max_problem}}
We provide two lemmas about Kronecker products before we prove Lemma \ref{lemma:MCCA_reformulate_max_problem}.
\begin{lemma}
\label{lemma:mixed_product_property}
\textit{For matrices $\mat{A}, \mat{B}, \mat{C}$, and $\mat{D}$ such that matrix products $\mat{AC}$ and $\mat{BD}$ can be calculated,}
\begin{align*}
(\mat{A} \otimes \mat{B})(\mat{C} \otimes \mat{D}) = \mat{AC} \otimes \mat{BD}.
\end{align*}
\end{lemma}

\begin{lemma}
\label{lemma:spectrum_property}
    \textit{For square matrices $\mat{A}$ and $\mat{B}$,}
    \begin{align*}
        \tr (\mat{A} \otimes \mat{B}) = \tr (\mat{A}) \tr (\mat{B}).
    \end{align*}
\end{lemma}
These lemmas are known as the mixed-product property and the spectrum property, respectively; see \cite{Harville_book} for detailed proofs.

\noindent \textbf{\textit{Proof of Lemma \ref{lemma:MCCA_reformulate_max_problem}}:}

For the maximization problem (\ref{eq:CCA_max_problem_rplc_ast}), move the summation over index $g$ out of the $\tr(\cdot)$ and replace $\mat{S}_{(g)}^*$ and $\mat{V}^*$ with $\mat{S}_{(g)}^{(1)} \otimes \mat{S}_{(g)}^{(2)} \otimes \cdots \otimes \mat{S}_{(g)}^{(M)}$ and $\mat{V}^{(1)} \otimes \mat{V}^{(2)} \otimes \cdots \otimes \mat{V}^{(M)}$, respectively. 
Then
\begin{align*}
    \max_{\substack{ \mat{V}^{(k)} \\ k=1,2,\dots, M }}  \sum_{g=1}^G \tr \left\{ \trans{{\qty(\mat{V}^{(1)} \otimes \cdots \otimes \mat{V}^{(M)})}} \qty(\mat{S}_{(g)}^{(1)} \otimes \cdots \otimes \mat{S}_{(g)}^{(M)}) \qty(\mat{V}^{(1)} \otimes \cdots \otimes \mat{V}^{(M)}) \right.\\
 \left. \trans{{\qty(\mat{V}^{(1)} \otimes \cdots \otimes \mat{V}^{(M)})}} \qty(\mat{S}_{(g)}^{(1)} \otimes \cdots \otimes \mat{S}_{(g)}^{(M)})
 \qty(\mat{V}^{(1)} \otimes \cdots \otimes \mat{V}^{(M)}) \right\},\\
 \qq{s.t.} \trans{{\mat{V}^{(k)}}} \mat{V}^{(k)} = \mat{I}_{R_k}.
\end{align*}
By Lemmas \ref{lemma:mixed_product_property} and \ref{lemma:spectrum_property}, we have
\begin{align*}
    &\max_{\substack{ {\mat{V}^{(k)}}^\top \mat{V}^{(k)} = \mat{I}_{R_k}  \\ k=1,2,\dots, M }}  \sum_{g=1}^G \tr \qty{ \qty( {\mat{V}^{(1)}}^\top \mat{S}_{(g)}^{(1)} \mat{V}^{(1)} {\mat{V}^{(1)}}^\top \mat{S}_{(g)}^{(1)} \mat{V}^{(1)} ) \cdots \qty( {\mat{V}^{(M)}}^\top \mat{S}_{(g)}^{(M)} \mat{V}^{(M)}{\mat{V}^{(M)}}^\top \mat{S}_{(g)}^{(M)} \mat{V}^{(M)} ) }\\
    =&\max_{\substack{ {\mat{V}^{(k)}}^\top \mat{V}^{(k)} = \mat{I}_{R_k}  \\ k=1,2,\dots, M }} \sum_{g=1}^G  \prod_{k=1}^{M} \tr \qty{  \trans{{\mat{V}^{(k)}}}  \mat{S}_{(g)}^{(k)} \mat{V}^{(k)} \trans{{\mat{V}^{(k)}}} \mat{S}_{(g)}^{(k)} \mat{V}^{(k)} }.
\end{align*}
This leads to the maximization problem in Lemma \ref{lemma:MCCA_reformulate_max_problem}.
\begin{align*}
    \tag*{$\qed$}
\end{align*}
\section{Proof of Theorem \ref{theorem:upper_and_lower_bound_max}}
Theorem \ref{theorem:upper_and_lower_bound_max} can be easily shown from the following lemma.
\begin{lemma}
    \label{lemma:general_bound}
    \textit{Consider the maximization problem}
    \begin{align}
    \max_{\mat{V}^{(k)}} f_k'({\mat{V}^{(k)}}) = \max_{\mat{V}^{(k)}} \tr\qty{ {\mat{V}^{(k)}}^\top \qty(\sum_{g=1}^G w_{(g)}^{(-k)} \mat{S}_{(g)}^{(k)} \mat{S}_{(g)}^{(k)} ) \mat{V}^{(k)} }.
    \end{align}
    \textit{Let $M^{(k)} = \tr \qty{\sum_{g=1}^G w_{(g)}^{(-k)} \mat{S}_{(g)}^{(k)} \mat{S}_{(g)}^{(k)} }$. Then}
    \begin{align*}
        \frac{f_k'(\mat{V}^{(k)})^2}{ M^{(k)}} \leq f_k(\mat{V}^{(k)}) \leq f_k'(\mat{V}^{(k)}).
    \end{align*}
\end{lemma}

\noindent \textbf{\textit{Proof of Lemma \ref{lemma:general_bound}}:}

First, we prove $f_k(\mat{V}^{(k)}) \leq f_k'(\mat{V}^{(k)})$. 
For any orthogonal matrix $\mat{V}^{(k)} \in \mathbb{R}^{P_k \times R_k}$, we can always find an orthogonal matrix $\mat{V}^{(k)}_\bot \in \mathbb{R}^{P_k \times (P_k - R_k)}$ that satisfies $\mat{V}^{(k)\top} \mat{V}^{(k)}_\bot = \bm O$. 
Then the equation $\mat{V}^{(k)} {\mat{V}^{(k)}}^\top + \mat{V}^{(k)}_\bot {\mat{V}^{(k)}_\bot}^\top = \mat{I}_{P_k}$ holds.
By definition,
\begin{align*}
    f_k(\mat{V}^{(k)}) &= \tr \qty{ {\mat{V}^{(k)}}^\top \qty( \sum_{g=1}^G w_{(g)}^{(-k)} \mat{S}_{(g)}^{(k)} \mat{V}^{(k)} {\mat{V}^{(k)}}^\top \mat{S}_{(g)}^{(k)} ) \mat{V}^{(k)} }\\
    &\leq \tr \qty{ {\mat{V}^{(k)}}^\top \qty( \sum_{g=1}^G w_{(g)}^{(-k)} \mat{S}_{(g)}^{(k)} \qty( \mat{V}^{(k)} {\mat{V}^{(k)}}^\top + \mat{V}^{(k)}_\bot {\mat{V}^{(k)}_\bot}^\top ) \mat{S}_{(g)}^{(k)} ) \mat{V}^{(k)} }\\
    &= \tr \qty{ {\mat{V}^{(k)}}^\top \qty( \sum_{g=1}^G w_{(g)}^{(-k)} \mat{S}_{(g)}^{(k)} \mat{S}_{(g)}^{(k)} ) \mat{V}^{(k)} }\\
    &= f_k'({\mat{V}^{(k)}}).
\end{align*}
Thus, we have obtained $f_k(\mat{V}^{(k)}) \leq f_k'({\mat{V}^{(k)}})$.

Next, we prove $\frac{f_k'(\mat{V}^{(k)})^2}{ M^{(k)}} \leq f_k(\mat{V}^{(k)})$.
We define the following block matrices:
\begin{align*}
    \mat{A} &= \qty[\sqrt{w_{(1)}^{(-k)}}{\mat{S}_{(1)}^{(k)}}^{\frac{1}{2}} \mat{V}^{(k)} {\mat{V}^{(k)}}^\top {\mat{S}_{(1)}^{(k)}}^{\frac{1}{2}}, \dots, \sqrt{ w_{(G)}^{(-k)} }{\mat{S}_{(G)}^{(k)}}^{\frac{1}{2}} \mat{V}^{(k)} {\mat{V}^{(k)}}^\top {\mat{S}_{(G)}^{(k)}}^{\frac{1}{2}} ],\\
    \mat{B} &= \qty[\sqrt{ w_{(1)}^{(-k)} } \mat{S}_{(1)}^{(k)}, \dots, \sqrt{ w_{(G)}^{(-k)} }\mat{S}_{(G)}^{(k)}]
\end{align*}
Note that since $\mat{S}_{(g)}^{(k)}$ is a symmetric positive definite matrix, $\mat{S}_{(g)}^{(k)}$ can be decomposed to ${\mat{S}_{(g)}^{(k)}}^{\frac{1}{2}}{\mat{S}_{(g)}^{(k)}}^{\frac{1}{2}}$.
We calculate the traces of $\mat{A}\mat{A}$, $\mat{A}\mat{B}$, and $\mat{B}\mat{B}$, respectively:
\begin{align*}
    \tr \qty(\mat{A}\mat{A}) &= \sum_{g=1}^G w_{(g)}^{(-k)} \tr \qty( {\mat{S}_{(g)}^{(k)}}^{\frac{1}{2}} \mat{V}^{(k)} {\mat{V}^{(k)}}^\top {\mat{S}_{(g)}^{(k)}}^{\frac{1}{2}} {\mat{S}_{(g)}^{(k)}}^{\frac{1}{2}} \mat{V}^{(k)} {\mat{V}^{(k)}}^\top {\mat{S}_{(g)}^{(k)}}^{\frac{1}{2}})\\
    &=\sum_{g=1}^G w_{(g)}^{(-k)} \tr  \qty{ {\mat{V}^{(k)}}^\top \mat{S}_{(g)}^{(k)} \mat{V}^{(k)} {\mat{V}^{(k)}}^\top \mat{S}_{(g)}^{(k)} \mat{V}^{(k)} }\\
    &= \tr \qty{ {\mat{V}^{(k)}}^\top \qty( \sum_{g=1}^G w_{(g)}^{(-k)} \mat{S}_{(g)}^{(k)} \mat{V}^{(k)} {\mat{V}^{(k)}}^\top \mat{S}_{(g)}^{(k)} ) \mat{V}^{(k)} }\\
    &= f_k(\mat{V}^{(k)}),\\ \\
    \tr \qty(\mat{A}\mat{B}) &= \sum_{g=1}^G w_{(g)}^{(-k)} \tr \qty( {\mat{S}_{(g)}^{(k)}}^{\frac{1}{2}} \mat{V}^{(k)} {\mat{V}^{(k)}}^\top {\mat{S}_{(g)}^{(k)}}^{\frac{1}{2}} \mat{S}_{(g)}^{(k)})\\
    &= \sum_{g=1}^G w_{(g)}^{(-k)} \tr \qty( {\mat{S}_{(g)}^{(k)}}^{\frac{1}{2}} \mat{V}^{(k)} {\mat{V}^{(k)}}^\top {\mat{S}_{(g)}^{(k)}}^{\frac{1}{2}} {\mat{S}_{(g)}^{(k)}}^{\frac{1}{2}} {\mat{S}_{(g)}^{(k)}}^{\frac{1}{2}} )\\
     &=\sum_{g=1}^G w_{(g)}^{(-k)} \tr  \qty{ {\mat{V}^{(k)}}^\top \mat{S}_{(g)}^{(k)} \mat{S}_{(g)}^{(k)} \mat{V}^{(k)} }\\
    &= \tr \qty{ {\mat{V}^{(k)}}^\top \qty( \sum_{g=1}^G w_{(g)}^{(-k)} \mat{S}_{(g)}^{(k)} \mat{S}_{(g)}^{(k)} ) \mat{V}^{(k)} }\\
    &= f_k'(\mat{V}^{(k)}),\\ \\
    \tr \qty(\mat{B}\mat{B}) &= \tr \qty( \sum_{g=1}^G w_{(g)}^{(-k)} \mat{S}_{(g)}^{(k)} \mat{S}_{(g)}^{(k)} ) = M^{(k)}.
\end{align*}
From the Cauchy--Schwarz inequality, we have
\begin{align*}
    f_k(\mat{V}^{(k)}) M^{(k)} = \tr \qty(\mat{A}\mat{A}) \tr \qty(\mat{B}\mat{B}) \geq \qty{\tr \qty(\mat{A}\mat{B})}^2 = f_k'(\mat{V}^{(k)})^2.
\end{align*}
By dividing both sides of the inequality by $M^{(k)}$, we obtain $\frac{f_k'(\mat{V}^{(k)})^2}{ M^{(k)}} \leq f_k(\mat{V}^{(k)})$. 
This completes the proof.
\begin{equation*}
    \tag*{$\qed$}
\end{equation*}

\noindent \textbf{\textit{Proof of Theorem \ref{theorem:upper_and_lower_bound_max}}:}

Let $f_k'^{\max}$ be the global maximum of $f_k'(\mat{V}^{(k)})$ and $\mat{V}_0^{(k)} = \argmax_{\mat{V}^{(k)}} f_k'(\mat{V}^{(k)})$.
From Lemma \ref{lemma:general_bound} and the definition of $\alpha^{(k)}$, we have
\begin{align*}
    \alpha^{(k)} f_k'^{\max} = \frac{f_k'(\mat{V}_0^{(k)})^2}{ M^{(k)}} \leq f_k(\mat{V}_0^{(k)}).
\end{align*}
Let $f_k^{\max}$ be the global maximum of $f_k(\mat{V}^{(k)})$. It then holds that $f_k(\mat{V}_0^{(k)}) \leq f_k^{\max}$. Thus
\begin{align*}
     \alpha^{(k)} f_k'^{\max} \leq f_k^{\max}.
\end{align*}
Let $\mat{V}_{0^*}^{(k)} = \argmax_{\mat{V}^{(k)}} f_k(\mat{V}^{(k)})$. From Lemma \ref{lemma:general_bound}, we have
\begin{align*}
    f_k^{\max} = f_k(\mat{V}_{0^*}^{(k)}) \leq f_k'(\mat{V}_{0^*}^{(k)}).
\end{align*}
Since $f_k'(\mat{V}_{0^*}^{(k)}) \leq f_k'^{\max}$, we have
\begin{align*}
    f_k^{\max} \leq f_k'^{\max}.
\end{align*}
Hence, we have obtained $\alpha^{(k)} f_k'^{\max} \leq f_k^{\max} \leq f_k'^{\max}$.
\begin{equation*}
    \tag*{$\qed$}
\end{equation*}

\section{Proof of Theorem \ref{theorem:optimal_alpha}}
\noindent \textbf{\textit{Proof of Theorem \ref{theorem:optimal_alpha}}:}
By definition 
\begin{align}
        \alpha^{(k)} =  \frac{f_k'^{\max}}{M^{(k)}} =  \frac{ \tr \qty{ {{\mat{V}_0^{(k)}}}^\top  \qty(\sum_{g=1}^G w_{(g)}^{(-k)} \mat{S}_{(g)}^{(k)} \mat{S}_{(g)}^{(k)}) \mat{V}_0^{(k)} }}{\tr \qty{\sum_{g=1}^G w_{(g)}^{(-k)} \mat{S}_{(g)}^{(k)} \mat{S}_{(g)}^{(k)} }}.
\end{align}
By using the eigenvalue representation, we can rewrite the numerator of $\alpha^{(k)}$ as follows:
\begin{align*}
     f_k'^{\max} = \sum_{g=1}^G w_{(g)}^{(-k)} \sum_{i=1}^{R_k} {\lambda_{(g)i}^{(k)}}.
\end{align*}
On the other hand, the denominator of $\alpha^{(k)}$ can be represented as the sum of eigenvalues as follows:
\begin{align*}
    M^{(k)} =  \sum_{g=1}^G w_{(g)}^{(-k)} \sum_{i=1}^{P_k} {\lambda_{(g)i}^{(k)}}.
\end{align*}
Thus, we can transform $\alpha^{(k)}$ as follows:
\begin{align*}
    \alpha^{(k)} = \frac{ \sum_{g=1}^G w_{(g)}^{(-k)} \sum_{i=1}^{R_k} {\lambda_{(g)i}^{(k)}}}{ \sum_{g=1}^G w_{(g)}^{(-k)} \sum_{i=1}^{P_k} {\lambda_{(g)i}^{(k)}}}.
\end{align*}
When we set
\begin{align*}
    \bm{\lambda}_0^{(k)} &= \qty[ \sum_{i=R_k + 1}^{P_k} {\lambda_{(1)i}^{(k)}}, \sum_{i=R_k + 1}^{P_k} {\lambda_{(2)i}^{(k)}}, \dots, \sum_{i=R_k + 1}^{P_k} {\lambda_{(G)i}^{(k)}} ]^\top,\\
    \bm{\lambda}_1^{(k)} &= \qty[ \sum_{i=1}^{P_k} {\lambda_{(1)i}^{(k)}}, \sum_{i=1}^{P_k} {\lambda_{(2)i}^{(k)}}, \dots, \sum_{i=1}^{P_k} {\lambda_{(G)i}^{(k)}} ]^\top,\\
    \bm{w}^{(k)} &= \qty[ w_{(1)}^{(-k)}, w_{(2)}^{(-k)}, \dots, w_{(G)}^{(-k)} ]^\top,
\end{align*}
we can reformulate $\alpha^{(k)}$ as
\begin{align*}
    \alpha^{(k)} = \frac{\qty(\bm{\lambda}_1^{(k)} - \bm{\lambda}_0^{(k)})^\top \bm{w}^{(k)} }{{\bm{\lambda}_1^{(k)}}^\top \bm{w}^{(k)} }.
\end{align*}
Thus, we obtain the following maximization problem:
\begin{align*}
    \max_{\bm{w}^{(k)}}  \frac{\qty(\bm{\lambda}_1^{(k)} - \bm{\lambda}_0^{(k)})^\top \bm{w}^{(k)} }{{\bm{\lambda}_1^{(k)}}^\top \bm{w}^{(k)} }, \qq{s.t.} \bm{w}^{(k)} > \bm{0} .
\end{align*}
Note that the constraints can be obtained by the definition of $\bm{w}^{(k)}$.
In addition, this maximization problem can be reformulated as
\begin{align*}
    \max_{\bm{w}^{(k)}}  \frac{\qty(\bm{\lambda}_1^{(k)} - \bm{\lambda}_0^{(k)})^\top \bm{w}^{(k)} }{{\bm{\lambda}_1^{(k)}}^\top \bm{w}^{(k)} } &= \max_{\bm{w}^{(k)}} 1 -  \frac{{\bm{\lambda}_0^{(k)}}^\top \bm{w}^{(k)} }{{\bm{\lambda}_1^{(k)}}^\top \bm{w}^{(k)} }\\
    &= \min_{\bm{w}^{(k)}} \frac{{\bm{\lambda}_0^{(k)}}^\top \bm{w}^{(k)} }{{\bm{\lambda}_1^{(k)}}^\top \bm{w}^{(k)} }.
\end{align*}
Since ${\bm{\lambda}_0^{(k)}}^\top \bm{w}^{(k)} / {\bm{\lambda}_1^{(k)}}^\top \bm{w}^{(k)}$ is non-negative, solving the optimization problem for the squared function of the objective function maintains generality. 
Thus, we can consider the following minimization problem:
\begin{align*}
   \min_{\bm{w}^{(k)}} \frac{{\bm{w}^{(k)}}^\top \bm{\lambda}_0^{(k)} {\bm{\lambda}_0^{(k)}}^\top \bm{w}^{(k)} }{ {\bm{w}^{(k)}}^\top \bm{\lambda}_1^{(k)} {\bm{\lambda}_1^{(k)}}^\top \bm{w}^{(k)} }, \qq{s.t.} \bm{w}^{(k)} > \bm{0}.
\end{align*}
Additionally, from the invariance under multiplication of $\bm{w}^{(k)}$ by a constant, we obtain the following objective function of the quadratic programming problem.
\begin{align*}
    \min_{\bm{w}^{(k)}} {\bm{w}^{(k)}}^\top \bm{\lambda}_0^{(k)} {\bm{\lambda}_0^{(k)}}^\top \bm{w}^{(k)}, \qq{s.t.} \bm{w}^{(k)} > \bm{0},\ {\bm{w}^{(k)}}^\top \bm{\lambda}_1^{(k)} {\bm{\lambda}_1^{(k)}}^\top \bm{w}^{(k)} = 1.
\end{align*}
The proof is complete.
\begin{equation*}
    \tag*{$\qed$}
\end{equation*}

\section{Proof of Theorem \ref{theorem:update_guarantee_increase}}
\noindent \textbf{\textit{Proof of Theorem  \ref{theorem:update_guarantee_increase}}:}
We define the following block matrices:
\begin{align*}
    \mat{A}_s = \qty[\sqrt{w_{(1)}^{(-k)}}{\mat{S}_{(1)}^{(k)}}^{\frac{1}{2}} \mat{V}_s^{(k)} {\mat{V}_s^{(k)}}^\top {\mat{S}_{(1)}^{(k)}}^{\frac{1}{2}}, \dots, \sqrt{ w_{(G)}^{(-k)} }{\mat{S}_{(G)}^{(k)}}^{\frac{1}{2}} \mat{V}_s^{(k)} {\mat{V}_s^{(k)}}^\top {\mat{S}_{(G)}^{(k)}}^{\frac{1}{2}} ].
\end{align*}
Here, we calculate the traces of $\mat{A}_s \mat{A}_s$, $\mat{A}_s \mat{A}_{s+1}$, and $\mat{A}_{s+1} \mat{A}_{s+1}$.
The calculations of $\tr \qty(\mat{A}_s \mat{A}_s)$ and $\tr \qty(\mat{A}_{s+1} \mat{A}_{s+1})$ are the same as that of $\tr \qty(\mat{A} \mat{A})$ by replacing $\mat{V}^{(k)}$ with $\mat{V}_s^{(k)}$ and $\mat{V}^{(k)}$ with $\mat{V}_{s+1}^{(k)}$, respectively, in Lemma \ref{lemma:general_bound}.
Thus, we obtain
\begin{align*}
    \tr \qty(\mat{A}_s \mat{A}_s) &= f_k(\mat{V}_s^{(k)}),\\ \\
    \tr \qty(\mat{A}_s \mat{A}_{s+1}) &= \sum_{g=1}^G w_{(g)}^{(-k)} \tr \qty( {\mat{S}_{(g)}^{(k)}}^{\frac{1}{2}} \mat{V}_s^{(k)} {\mat{V}_s^{(k)}}^\top {\mat{S}_{(g)}^{(k)}}^{\frac{1}{2}} {\mat{S}_{(g)}^{(k)}}^{\frac{1}{2}} \mat{V}_{s+1}^{(k)} {\mat{V}_{s+1}^{(k)}}^\top {\mat{S}_{(g)}^{(k)}}^{\frac{1}{2}})\\
    &=\sum_{g=1}^G w_{(g)}^{(-k)} \tr  \qty{ {\mat{V}_{s+1}^{(k)}}^\top \mat{S}_{(g)}^{(k)} \mat{V}_s^{(k)} {\mat{V}_s^{(k)}}^\top \mat{S}_{(g)}^{(k)} \mat{V}_{s+1}^{(k)} }\\
     &=\tr \qty{{\mat{V}_{s+1}^{(k)}}^\top \qty( \sum_{g=1}^G w_{(g)}^{(-k)}  \mat{S}_{(g)}^{(k)} \mat{V}_s^{(k)} {\mat{V}_s^{(k)}}^\top \mat{S}_{(g)}^{(k)} ) \mat{V}_{s+1}^{(k)} },\\ \\
    \tr \qty(\mat{A}_{s+1} \mat{A}_{s+1}) &= f_k(\mat{V}_{s+1}^{(k)}).\\ \\
\end{align*}
Since $\mat{V}_{s+1}^{(k)} = \argmax_{\mat{V}^{(k)}} \tr \qty{ {\mat{V}^{(k)}}^\top \qty( \sum_{g=1}^G w_{(g)}^{(-k)}  \mat{S}_{(g)}^{(k)} \mat{V}_s^{(k)} {\mat{V}_s^{(k)}}^\top \mat{S}_{(g)}^{(k)} ) \mat{V}^{(k)} }$, we have 
\begin{align*}
    f_k(\mat{V}_s^{(k)}) &= \tr  \qty{ {\mat{V}_s^{(k)}}^\top \qty( \sum_{g=1}^G w_{(g)}^{(-k)}  \mat{S}_{(g)}^{(k)} \mat{V}_s^{(k)} {\mat{V}_s^{(k)}}^\top \mat{S}_{(g)}^{(k)} ) \mat{V}_s^{(k)} }\\
    &\leq \tr \qty{{\mat{V}_{s+1}^{(k)}}^\top \qty( \sum_{g=1}^G w_{(g)}^{(-k)}  \mat{S}_{(g)}^{(k)} \mat{V}_s^{(k)} {\mat{V}_s^{(k)}}^\top \mat{S}_{(g)}^{(k)} ) \mat{V}_{s+1}^{(k)} }\\
    &= \tr \qty(\mat{A}_s \mat{A}_{s+1}).
\end{align*}
From the positivity of both sides of the inequality, it holds that
\begin{align*}
    f_k(\mat{V}_s^{(k)})^2 \leq  \qty[\tr \qty(\mat{A}_s \mat{A}_{s+1})]^2.
\end{align*}
In addition, from the Cauchy--Schwarz inequality, we have
\begin{align*}
    f_k(\mat{V}_s^{(k)}) f_k(\mat{V}_{s+1}^{(k)}) &= \tr \qty(\mat{A}_s \mat{A}_s) \tr \qty(\mat{A}_{s+1} \mat{A}_{s+1})\\
    &\geq \qty[\tr \qty(\mat{A}_s \mat{A}_{s+1})]^2.
\end{align*}
Thus,
\begin{align*}
    f_k(\mat{V}_s^{(k)}) f_k(\mat{V}_{s+1}^{(k)}) \geq \qty[\tr \qty(\mat{A}_s \mat{A}_{s+1})]^2 \geq f_k(\mat{V}_s^{(k)})^2.
\end{align*}
Thus, we have obtained $f_k(\mat{V}_s^{(k)})^2 \leq f_k(\mat{V}_s^{(k)}) f_k(\mat{V}_{s+1}^{(k)})$.
By dividing both sides of the inequality by $ f_k(\mat{V}_s^{(k)})$, we obtain the relation $f_k(\mat{V}_s^{(k)}) \leq  f_k(\mat{V}_{s+1}^{(k)})$.
\begin{equation*}
    \tag*{$\qed$}
\end{equation*}

\section*{Acknowledgments}

S. K. was supported by JSPS KAKENHI Grant Numbers JP19K11854 and JP20H02227, and MEXT KAKENHI Grant Numbers JP16H06429, JP16K21723, and JP16H06430.

\bibliographystyle{apalike}
\bibliography{mcca}

\end{document}